\documentclass[11pt,a4paper]{article}
\usepackage[utf8]{inputenc}
\usepackage{graphicx}
\usepackage{xcolor}
\definecolor{esablue}{RGB}{0,57,158}
\definecolor{esalightblue}{RGB}{0,155,219}
\definecolor{esared}{RGB}{150,1,54}
\usepackage[hidelinks]{hyperref}
\hypersetup{
  colorlinks   = true, 
  urlcolor     = esalightblue, 
  linkcolor    = esablue, 
  citecolor   = esablue 
}
\usepackage{amsmath}
\usepackage{amssymb}

\newcommand{\etal}{\textit{et al.}}

\usepackage[british]{babel}
\usepackage{csquotes}

\usepackage[textsize=tiny,
backgroundcolor=white,
textcolor=blue,
linecolor=blue,
bordercolor=blue]{todonotes}
\usepackage[capitalise]{cleveref}

\usepackage{caption}
\usepackage{subcaption}
\usepackage{tabularx}
\usepackage{multirow}
\usepackage{tablefootnote}
\usepackage[flushleft]{threeparttable}

\usepackage{booktabs}
\renewcommand{\arraystretch}{1.2}

\usepackage{amsfonts}
\newcommand{\deriv}[1]{\frac{\mathrm{d}}{\mathrm{d}#1}}

\usepackage[
backend=biber,
citestyle=numeric-comp,
bibstyle=numeric,
minbibnames=2,%
maxbibnames=3,%
maxcitenames=3,%
style=numeric,
autocite=plain,
sortcites=true,
uniquename=init,
giveninits=true,
isbn=true,
doi=true,
url=true,%
natbib=true,%
urldate=short,
sorting=none,
arxiv=abs,
dateabbrev=false
]{biblatex}
\renewbibmacro{in:}{}

\AtEveryBibitem{%
	\ifentrytype{inproceedings}{
		\clearlist{publisher}
		\clearfield{volume}
		\clearfield{pages}
	}{}
	\clearlist{address}
	\clearlist{location}
    \ifentrytype{misc}{}{
        \clearfield{month}
    	\clearfield{day}
        \clearfield{url}
        \clearfield{urldate}
    }
	\clearfield{series}
	\clearfield{primaryclass}
	\clearfield{eprintclass}
}

\title{Neuromorphic Computing and Sensing in Space}
\author{Dario Izzo\thanks{Advanced Concepts Team (ACT), European Space Research \& Technology Centre (ESTEC), Keplerlaan 1, 2200 AG Noordwijk (Netherland)} \thanks{All authors have contributed equally to this work.}, 
Alexander Hadjiivanov$^*$$^\dagger$, 
Dominik Dold$^*$$^\dagger$ \\ 
Gabriele Meoni$^*$$^\dagger$\thanks{$\Phi$-lab,  European Space Research Institute (ESRIN), Via Galileo Galilei, 1, 00044 Frascati RM (Italy)}, Emmanuel Blazquez$^*$$^\dagger$}
\date{}

\begin{document}

\maketitle
\begingroup
\hypersetup{linkcolor=black}
\endgroup
\section{Introduction}

The term ``neuromorphic'' refers to systems that are closely resembling the architecture and/or the dynamics of biological neural networks~\cite{frenkel2021bottom, roy2019towards, indiveri2011frontiers}.
Typical examples would be novel computer chips designed to mimic the architecture of a biological brain, or sensors that get inspiration from, e.g., the visual or olfactory systems in insects and mammals to acquire information about the environment.
This approach is not without ambition as it promises to enable engineered devices able to reproduce the level of performance observed in biological organisms -- the main immediate advantage being the efficient use of scarce resources, which translates into low power requirements.
Nowadays, the neuromorphic approach is mostly investigated at two levels (i) algorithmic and (ii) hardware. On the algorithmic level, it leverages spike-based processing and training \cite{roy2019towards} to build novel machine learning pipelines able to process data efficiently. 
At the hardware level, the neuromorphic approach is pursued in designing novel analog and digital circuits and computer chips inspired by biological neural systems.
This results in novel sensing devices believed to produce particularly good candidates to emulate biological vision, as well as in the design of computer chips dedicated to efficiently implement the spike-based systems just introduced.
In fact, due to the discontinuous nature of spike-based communication and the temporal dynamics of spiking neurons, simulating the behaviour of a whole network of spiking neurons on conventional computer hardware is computationally -- and thus energy-wise -- very inefficient. This has now also a close relative in the field of artificial intelligence (AI) where Geoffrey Hinton has recently introduced the concept of ``Mortal Computation''\cite{hintonforward}: a form of computing where no separation between software and hardware exists. In ``Mortal Computation'', neural network solutions are uniquely tied to their underlying (analogue) hardware substrate, which Hinton argues (according to us correctly) might be the only way of obtaining large-scale neural networks that are energy-efficient -- a concept that is closely following the neuromorphic paradigm. 

The emphasis on low power and energy efficiency of neuromorphic devices is a perfect match for space applications. Spacecraft -- especially miniaturized ones -- have strict energy constraints as they need to operate in an environment which is scarce with resources and extremely hostile \cite{furano2020towards}. 
Numerous works have been investigating different energy-efficient solutions, especially leveraging commercial off-the-shelf (COTS) devices, aiming at optimizing model performance and energy usage trade-offs \cite{furano2020towards,UnibapSpaceCloud, PhiSat1, OPSSATOnboardAI, danielsen2021self} much less have investigated neuromorphic devices. Early work \cite{Medici2010Neuromorphic, Valette2009Neuromorphic} performed in 2010 at the Advanced Concepts Team (ACT) suggested considering a neuromorphic approach for onboard spacecraft applications.
Focusing on optic flow detection \cite{valette2010biomimetic, izzo2011constant}, these preliminary works showed the possibility to safely land a spacecraft on an unknown planetary surface, assuming a neuromorphic approach to sensing based on the Elementary Motion Detector (EMD) \cite{frye2015elementary}, a device inspired by the visual system in flying insects. 
More recently, with the availability of new neuromorphic sensors such as the Dynamic Vision Sensor \cite{BrandliBernerYangEtAl_2014_240180130,TaverniPaulMoeysLiEtAl_2018_FrontBackIlluminated} and chips such as 
Loihi \cite{davies2018loihi}, TrueNorth \cite{akopyan2015truenorth}, Akida and others, the interest on neuromorphic architectures for spacecraft missions grew considerably.

Another significant potential advantage of neuromorphic hardware for space applications concerns radiation. Earth's atmosphere and magnetic field protects us from a lot of the cosmic radiation, but this poses a considerable problem even in relatively low orbits. While in many instances it can damage the actual hardware, radiation can also interfere with its operation (for instance, by flipping bits in memory), leading to software failure. Neuromorphic hardware can potentially mitigate these issues since, apart from intermittent spikes, it is in fact mostly silent. \cite{YeLiuTaggartEtAl_2019_EvaluationRadiationEffects}. 

In addition to processors, also event-based cameras have contributed to the growing interest in the field of neuromorphic engineering. Event-based vision sensors are well equipped for operation in space: they have a very high dynamic range (on the order of $120 dB$), respond only to moving segments in the visual field with very low latency and, most importantly, consume very little power due to their sparse output (on the order of $mW$). Naturally, these advantages come at a price -- for instance, very high noise in very dark environments and proportionally low fidelity for slow-moving objects in the visual field. 

In \cref{subsec: eventSensing}, we present several research lines that aim to harness these advantages and mitigate the downsides of event-based vision.
Before diving into neuromorphic sensing though, we will first give a brief introduction to neuromorphic algorithms (\cref{sec:snn,sec:learningSNN}) and neuromorphic hardware platforms (\cref{sec:nhw}) to then discuss past and current research mainly conducted by the ACT on evaluating the feasibility of a neuromorphic approach for onboard AI applications (\cref{sec:nmhwappl}).
We hope that this chapter will stimulate further research pursuing a neuromorphic approach to spacecraft onboard computation and sensing.

\section{Spiking neural networks}\label{sec:snn}
\begin{figure}[tb]
    \centering
    \includegraphics[width=\linewidth]{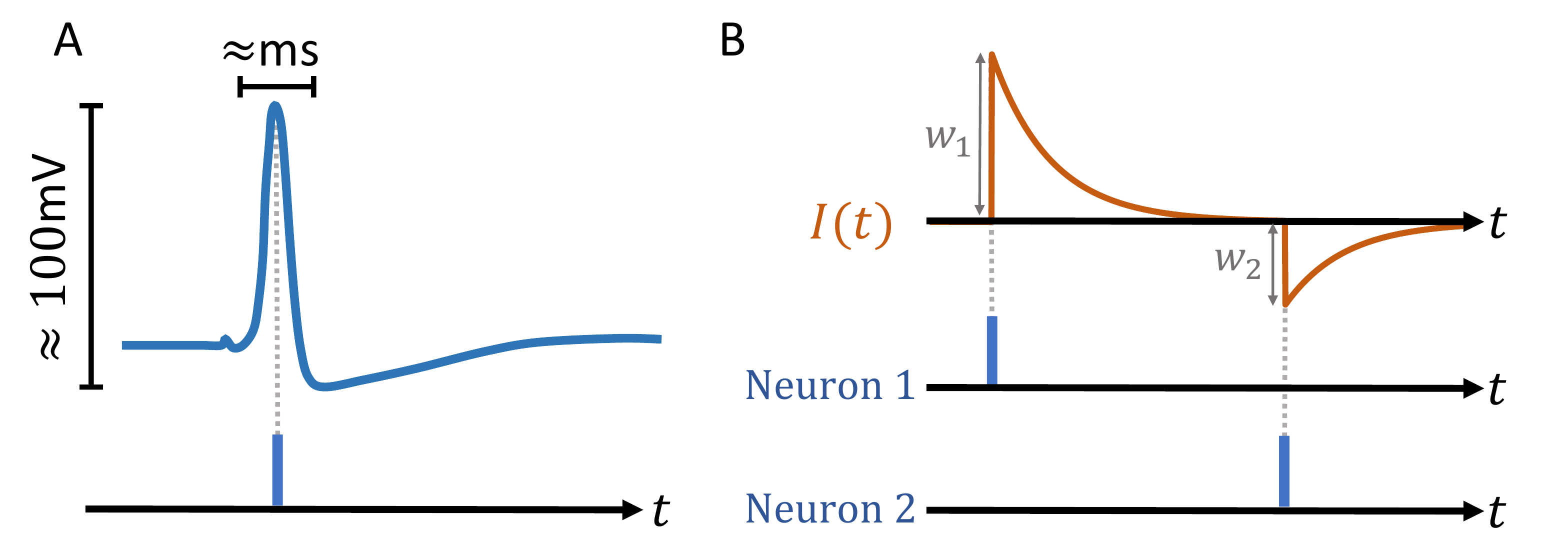}
    \caption{\textbf{(A)} Illustration of an action potential or sodium spike \cite{hodgkin1939action}, an electrical pulse that is sent through the axon, which acts as the `output cable' of a neuron, as a signal to other neurons. For computational purposes, it can be reduced to simply the time of its occurrence (bottom).
        \textbf{(B)} Presynaptic spikes trigger currents $I(t)$ at the postsynaptic neuron, with strength and direction depending on the interaction strength $w_i$ (commonly referred to as synaptic \textit{weight} in NN research).}\label{fig:action_potential}
\end{figure}

Arguably, the feature which is found most often in modern neuromorphic algorithms is spike-based communication.
In the mammalian brain, neurons communicate with each other using action potentials (`spikes' or `events') -- electrical pulses with a stereotypical shape where only the time at which the spike occurs carries information (\cref{fig:action_potential}A).
This realizes a highly sparse, and hence energy efficient, computing paradigm, as neurons only actively change their internal state when excited by an incoming spike, remaining passive otherwise.

\begin{figure}[ht]
    \centering
    \includegraphics[width=\linewidth]{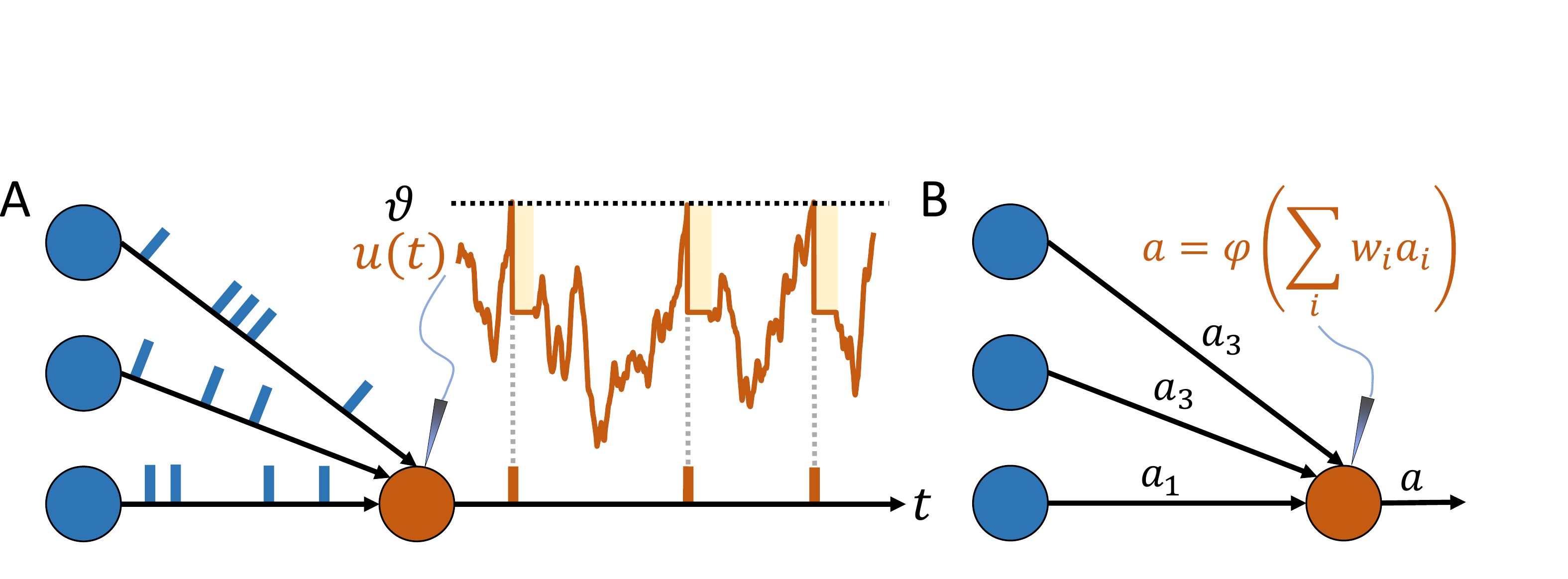}
    \caption{Comparison of SNNs and ANNs. \textbf{(A)} In a LIF-based SNN, presynaptic spikes (left, blue) lead to fluctuations in the membrane potential of the postsynaptic neuron (right, orange). When the membrane $u$ reaches a threshold value $\vartheta$, a spike is emitted and the membrane potential is clamped to a reset value for a period of time during which subsequent spiking is impossible (area highlighted in yellow).
        \textbf{(B)} In an ANN, real-valued activations $a_i$ are multiplied by weights $w_i$ and summed up at the postsynaptic neuron. The output $a$ of the postsynaptic neuron is a non-linear function $\varphi(\cdot)$ of this sum.}\label{fig:SNN_schematic}
\end{figure}

Mathematical models of spiking neurons differ substantially in terms of computational complexity and biological realism.
A widely used model (both in terms of algorithms and neuromorphic hardware) is the Leaky Integrate-and-Fire (LIF) neuron model, which adequately balances complexity and realism.
The LIF model represents a biological neuron as an RC-circuit\footnote{I.e., an electrical circuit with a resistor and capacitor coupled in parallel.} with capacitance $C$ and resistance $R$.
The capacitor maintains an electric voltage $u$ (the `membrane potential') and is charged by an electric current $I$ on a characteristic time scale $\tau = R \cdot C$ (the `membrane time constant'):
\begin{equation}
    \tau \deriv{t}u(t) = - u(t) + R \cdot I(t) \,. \label{eq:LIF}
\end{equation}
More specifically, $u(t)$ represents the potential difference between the inside and outside of the neural cell caused by different ion concentrations.
Via $R$, ions can `leak' through the membrane until an equilibrium potential is reached (here, the equilibrium potential is set to $0$ for simplicity).
The current $I(t)$ is caused by the spikes of source (`presynaptic') neurons (labelled with indices $i \in \mathbb{N}$ here), which connect to the target (`postsynaptic') neuron through synapses with interaction strengths $w_i \in \mathbb{R}$.
In the absence of spikes, $I(t)$ decays to $0$ and increases or decreases only whenever a presynaptic spike arrives, with strength and polarity depending on the value and sign of $w_i$ (\cref{fig:action_potential}B).
The postsynaptic neuron emits a spike at time $t_\mathrm{s}$ when the membrane potential crosses a threshold value\footnote{This threshold is not physically manifested in biological neurons, but it is a reasonably good approximation for modelling their response behaviour \cite{gerstner2009good}.} $\vartheta$, $u(t_\mathrm{s}) = \vartheta$ (\cref{fig:SNN_schematic}A). In biological neurons, the ability to spike is diminished for a short period of time (known as the `refractory period') immediately after spiking, as the mechanism responsible for creating action potentials has to recover first.
This is often modelled by clamping the membrane potential to a reset value for the refractory period (\cref{fig:SNN_schematic}A, yellow shaded areas).
Without the leak term $-u(t)$, the model is reduced to the Integrate-and-Fire (IF) neuron model.

The realism of the LIF model can be further increased by extending \cref{eq:LIF} with additional terms.
For instance, the Adaptive Exponential LIF model (AdEx) \cite{brette2005adaptive} further adds action potential shapes to $u(t)$ as well as spike-rate adaptation, where prolonged tonic spiking (i.e., permanent activity) of a neuron is preceded by a period of increased or decreased activity.
The temporal dynamics of AdEx neurons is capable of replicating a broad range of dynamics observed in biological neurons. In particular, it has been found that adaptation is employed by neurons for solving temporal tasks such as sequence prediction (see, e.g., \cite{bellec2018long,yin2021accurate}).

\begin{figure}[ht]
    \centering
    \includegraphics[width=\linewidth]{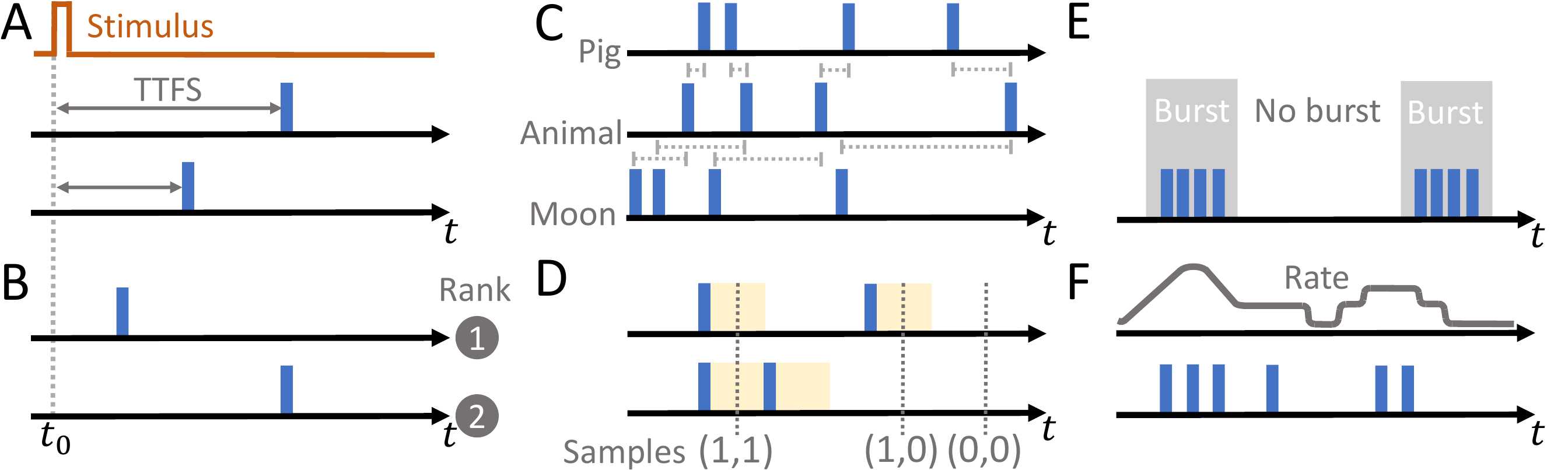}
    \caption{An illustration of spike-based encoding schemes. \textbf{(A)} TTFS encoding.
        \textbf{(B)} Rank order encoding with a population of two neurons.
        \textbf{(C)} Example of how concepts can be encoded and decoded (via similarity) in spike trains. Here, the spike train representing `Pig' and `Animal' are similar, while `Moon' and `Animal' are dissimilar. For clarity, the differences between spike times are shown in gray.
        \textbf{(D)} Encoding as random samples. At every point in time, a sample can be read out from the network, with neurons being in state '1' while refractory (yellow area) and in state '0' otherwise.
        \textbf{(E)} Encoding with bursts, short and intense periods of spike activity.
        \textbf{(F)} Rate-based encoding.}\label{fig:SNN_Coding}
\end{figure}

In principle, every architecture that exists for ANNs can be converted into a spiking neural network (SNN) by replacing artificial neurons with spiking neurons, e.g., LIF neurons.
However, ANNs and SNNs are different in two key aspects: (i) as discussed above, spiking neurons are characterised by a dynamic internal state, while artificial neurons possess no intrinsic state or dynamics, and therefore have no intrinsic `awareness' of time; and (ii) spiking neurons interact by triggering currents at specific times through spikes, while artificial neurons communicate directly with each other via real-valued activations (\cref{fig:SNN_schematic}B).
Even though the event-based nature of SNNs makes them potentially more energy efficient than ANNs, it is still an open question whether it also provides other advantages (e.g., in terms of performance, robustness to noise or training time).
A key challenge hereby is identifying how information can be efficiently encoded in the temporal domain of spikes \cite{zenke2021visualizing}.
A selection of proposed spike-based encoding schemes addressing this question is summarized in \cref{tab:SNNcoding} and illustrated in \cref{fig:SNN_Coding}.

\begin{table}[ht]
    \small
    \centering
    \resizebox{\textwidth}{!}{\begin{tabular}{p{0.2\linewidth}  p{0.8\linewidth}}
            \toprule
            \textbf{Encoding scheme}                                   & \textbf{Description}                \\ \midrule
            \midrule
            Time-To-First-Spike (TTFS)~\cite{thorpe1996speed} & Given a stimulus onset at time $t_0$, information is encoded in the time required for \textit{any} neuron in a population of neurons to emit its first spike (\cref{fig:SNN_Coding}A). This encoding scheme allows fast processing (i.e. low latency) and is highly energy efficient since each involved neuron spikes only once at the most in order to solve a given task.                                                                                   \\
            \midrule
            Rank order~\cite{thorpe1996speed}                 & Information is encoded by the order in which spikes occur in a population of neurons given the onset of a stimulus (\cref{fig:SNN_Coding}B). Thus, in contrast to TTFS, the exact spike time becomes irrelevant, but information can still be processed using only a low number of spikes depending on how many of the earliest spikes are used for encoding the input.                                                                                                                            \\
            \midrule
            Spike patterns                                    & Instead of only using single spikes to represent information, whole spike patterns of individual neurons or populations can be used \cite{gutig2006tempotron,dold2022neuro}. For instance, in \cite{dold2022relational}, symbolic information (e.g., abstract concepts) is represented by spike trains, and relatedness between concepts is encoded in the similarity of spike trains (\cref{fig:SNN_Coding}C). \\
            \midrule
            Sampling                                          & In neural sampling, neurons represent sampled values of binary random variables (refractory after spiking = $1$, non-refractory = $0$), allowing them to encode probability distributions (\cref{fig:SNN_Coding}D). This is suitable for approximating, e.g., sampling-based deep learning architectures such as restricted Boltzmann machines \cite{buesing2011neural,petrovici2016stochastic}.                      \\
            \midrule
            Bursts                                            & Information can be encoded in bursts, i.e., short periods of high spike activity (\cref{fig:SNN_Coding}E). For instance, in \cite{payeur2021burst}, bursts have been used to propagate information through SNNs that guides learning.                                                                                                                                                                          \\
            \midrule
            Rates                                             & An instantaneous spike rate obtained by averaging over spiking activity (either over time or populations) carries the information (\cref{fig:SNN_Coding}F). Since the spiking rate is a continuous variable, it encodes information in a similar way to ANNs.                                                                                                                                                    \\
            \bottomrule
        \end{tabular}}
    \caption{Different ways of encoding information using spikes.}
    \label{tab:SNNcoding}
\end{table}
In general, for machine learning applications it would be preferred to have a learning algorithm that automatically finds the optimal (combination of) encoding schemes.
Although training SNNs has been a daunting task for a long time, recent progress in terms of both theory and software infrastructure has enabled exactly such end-to-end learning for SNNs, opening novel opportunities for building highly efficient and powerful spike-based AI systems.





\section{Learning algorithms for SNNs}\label{sec:learningSNN}

\begin{figure}[ht]
    \centering
    \includegraphics[width=\linewidth]{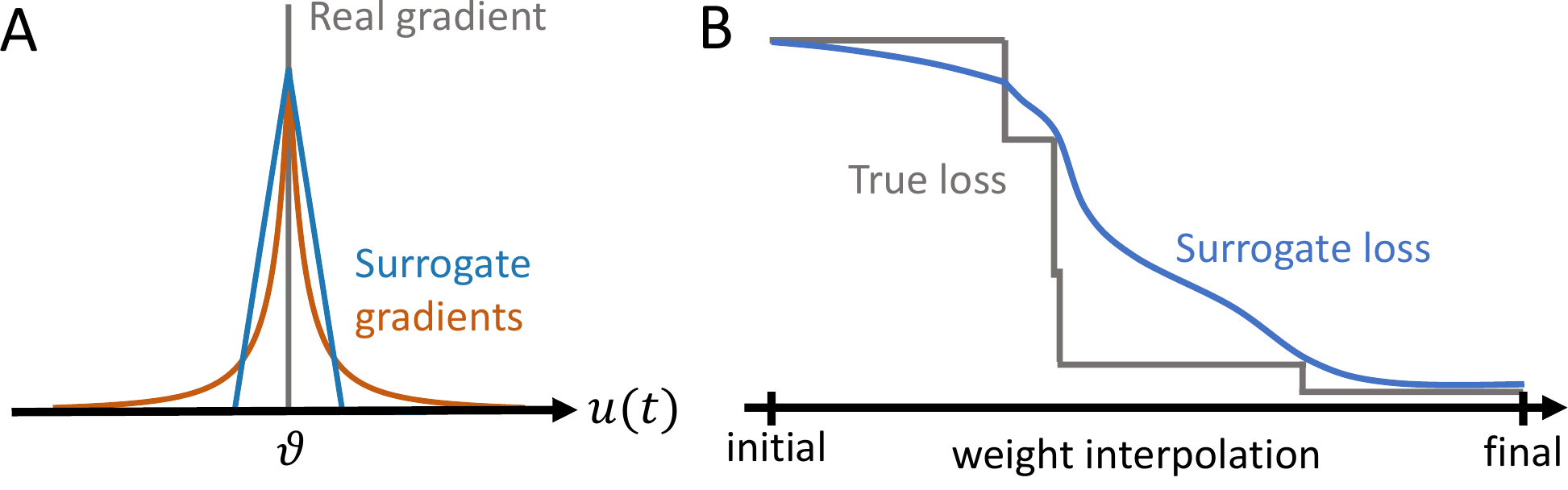}
    \caption{Illustration of the surrogate gradient method. \textbf{(A)} The gradient of the threshold function is a Dirac delta peak, which is infinite at $u = \theta$ and vanishes otherwise (gray). In the surrogate gradient approach, this gradient is replaced by a well-behaved function, such as a mirrored Lorentz function (orange) or a triangular function (blue). \textbf{(B)} In a SNN, the landscape of the loss function has discontinuities due to the spiking mechanism (gray). Especially long plateaus with constant loss inhibit learning. Using surrogate gradients `smoothens' the loss landscape (blue trace). Here, the loss landscape is illustrated by interpolating between two sets of weights of a SNN. Reproduced based on \cite{neftci2019surrogate}.}\label{fig:surrogate_gradient}
\end{figure}

One of the main enablers of the incredible success of deep learning in recent years is the error backpropagation algorithm (`backprop').
However, for a long time, this success did not pass over to SNNs -- mostly due to the threshold mechanism of spiking neurons that leads to vanishing gradients at all times except at the time of threshold crossing \cite{neftci2019surrogate} (cf. \cref{fig:surrogate_gradient} for an illustration of the concept of surrogate gradients, which present one solution to this problem).
Instead, to apply SNNs to a variety of machine learning applications, it was customary to convert the parameters of a trained ANN to a SNN \cite{pfeiffer2018deep}.
SNNs `trained' this way typically show a significant degradation in performance compared to the original ANN. In addition, the mapping promotes purely rate-based encoding in the SNNm which has been found to reduce the energy efficiency of SNNs \cite{davidson2021comparison,kucik2021investigating}.
An alternative approach is to use biologically inspired learning rules, such as Spike-Time-Dependent Plasticity (STDP) \cite{bi1998synaptic} or variants thereof, although they do not scale well beyond shallow networks\footnote{As of late, several biologically plausible learning rules that are applicable to deep neural architectures have been proposed \cite{illing2021local}. However, they lack the flexibility and theoretical guarantees of backprop, i.e., they are not proven to minimize a task-dependent (and customizable) loss function. Nevertheless, they present intriguing alternatives for training SNNs.}.

Recently, several approaches have been found for successfully unifying SNNs and backprop, which have become the \textit{de facto} state of the art for training SNNs on machine learning tasks and are currently being investigated at the ACT.
For instance, one approach is based on SNNs using neuron models where the TTFS is both analytically calculable and differentiable, allowing gradients to be calculated exactly without having to deal with the discrete nature of the threshold mechanism \cite{mostafa2017supervised,comsa2020temporal,kheradpisheh2020temporal,goltz2021fast}, although this method is limited to neurons that only spike once.

At the time of writing, the most widely used approach is the so-called ``Surrogate Gradient'' method, which is applicable to all kinds of spiking neuron models \cite{zenke2018superspike,neftci2019surrogate}.
Here, the gradient of the threshold function is replaced with a surrogate function that has non-zero parts for membrane potential values away from the threshold (\cref{fig:surrogate_gradient}A).
This `softening' of the spike threshold gradient function allows gradient information to flow continuously through the network (\cref{fig:surrogate_gradient}B), enabling end-to-end training of SNNs capable of utilizing the temporal domain of spikes to find, e.g., highly sparse and energy efficient solutions.
For instance, in \cite{zenke2021remarkable} SNNs are trained that reach a competitive classification accuracy ($\approx 1.7$\% test error) on the MNIST handwritten digits dataset \cite{lecun1998mnist} with, on average, only $10$ to $20$ spikes per inference.
Surrogate gradients can be used to optimise not only the weights but all parameters that influence the dynamic behaviour of spiking neurons, such as time constants and spike thresholds of individual neurons -- with potential benefits for the robustness and expressiveness of SNNs \cite{perez2021neural,yin2021accurate}.
A major downside of the surrogate gradient method is that currently there is no theoretical framework for choosing the shape of the surrogate gradient function.
Nevertheless, initial evidence suggests that the approach is relatively robust with respect to this choice \cite{zenke2021remarkable}.
In the future, this downside could be alleviated through exact derivations of gradient-based learning rules for SNNs (see, e.g., \cite{wunderlich2021event}).
Additional information on the current standards for training SNNs can be found in \cite{neftci2019surrogate,eshraghian2021training}.

To summarize, the surrogate gradient method enables end-to-end training of SNNs using error backpropagation.
Several open-source packages are available that standardize and ease the implementation and training of SNNs by utilising existing libraries for automatic differentiation.
Currently, four general SNN libraries based on pyTorch are being developed: Norse \cite{norse2021}, spikingjelly \cite{SpikingJelly}, snnTorch \cite{eshraghian2021training} and BindsNET \cite{bindsnet}.
The first three adopt both the workflow and class structure of pyTorch, effectively extending it to support SNNs.
The last one is geared towards developing machine learning algorithms that take inspiration from biology.
Even though it does not support gradient-based learning, it contains a larger variety of neuron models and biologically inspired learning rules than the other three packages.
In general, the emerging landscape of SNN libraries greatly reduces the development time of SNN-based algorithms, painting a promising picture for further exploration of the capabilities and potential benefits of SNNs.
In addition, it increases the accessibility of SNNs to researchers outside the field of neuromorphic computing and computational neuroscience by making it possible to seamlessly exchange or interweave ANNs and SNNs in a common framework. Thus, in the coming years we can expect to see an increasing number of contributions from the aerospace field attempting to incorporate spike-based algorithms and hardware onboard spacecraft.

\section{Existing hardware}\label{sec:nhw}

Although end-to-end training of SNNs is possible nowadays, exploring their capabilities remains challenging due to the computationally demanding nature of simulating the internal neuron dynamics of spiking neurons.
In contrast to artificial neurons, the dynamics of spiking neurons have to be solved using numerical solvers for ordinary differential equations, introducing a significant overhead.
Thus, on conventional hardware systems, the benefit of spike-based coding (in the absence of spikes, no active computations are performed) is not immediately apparent.

An emerging technology that is capable of harnessing the potential of spike-based information processing is neuromorphic hardware, which explores novel computing architectures and paradigms that closely emulate how the brain processes information. Standard computer architectures separate processing units and memory (storing data and instructions for the processor), introducing a bottleneck due to the constant flow of data between memory and processor (the so-called `von Neumann bottleneck’). Instead, neuromorphic hardware follows several design philosophies that can be found in the mammalian brain. 

\begin{itemize}
    \item{No separation between processing and memory. Thus, a processing unit (e.g., a neuron) can only access information that is locally available (e.g., synaptic weights or activity of other neurons that it is connected to);}
    \item{Large-scale parallel computing;}
    \item{Asynchronous instead of clocked computations;}
    \item{Event-based information processing (i.e., using spikes instead of continuous values);}
    \item{(Re)programming the chip consists of (re)mapping networks or adapting network parameters through learning;}
    \item{Time-continuous and locally constrained learning rules.}
\end{itemize}

Broadly speaking, neuromorphic hardware comes in two flavours: digital and analogue. Both designs are transistor-based (i.e., CMOS), but the transistors are used in different operating regimes. In simple terms, digital means calculating with discrete-valued bits (i.e., transistors take the states `on’ and `off’), while analogue means calculating with continuous-valued currents and voltages. Digital chips provide stable solutions that are closer to commonly used hardware chips. Analogue chips have to deal with noise (e.g., manufacturing noise of components, cross-talk between electrical components, temperature dependence) and have a longer development cycle, but provide intriguing advantages such as high energy efficiency and potentially accelerated (compared to biology) emulation of neuron and network dynamics. In addition to CMOS, memristor-based circuits\footnote{A memristor is a fundamental electrical element that displays a hysteretic resistance profile where the resistance depends on the recent history of current that has passed through the element.} are being developed that open novel opportunities for energy-efficient and adaptive neuromorphic devices.

Examples of digital neuromorphic systems are Loihi (1 and 2) (developed by Intel) \cite{davies2018loihi}, TrueNorth (developed by IBM)\cite{akopyan2015truenorth}, Akida (developed by Brainchip), SpiNNaker(1 and 2) \cite{furber2012overview,mayr2019spinnaker}, Darwin \cite{shen2016darwin} and Tianjic \cite{pei2019towards}. Examples of analogue chips are BrainScaleS (1 and 2) \cite{pehle2022brainscales} and Spikey \cite{pfeil2013six}, DYNAPs \cite{moradi2017scalable}, ROLLS \cite{qiao2015reconfigurable} and Neurogrid \cite{benjamin2014neurogrid}. A thorough review of state-of-the-art neuromorphic platforms at the time of writing can be found in \cite{frenkel2021bottom}.
Two of these platforms (Loihi by Intel and BrainScaleS-2 by the University of Heidelberg) are discussed in more detail below to further illustrate the difference between digital and analogue platforms.

\begin{figure}[ht]
    \centering
    \includegraphics[width=\linewidth]{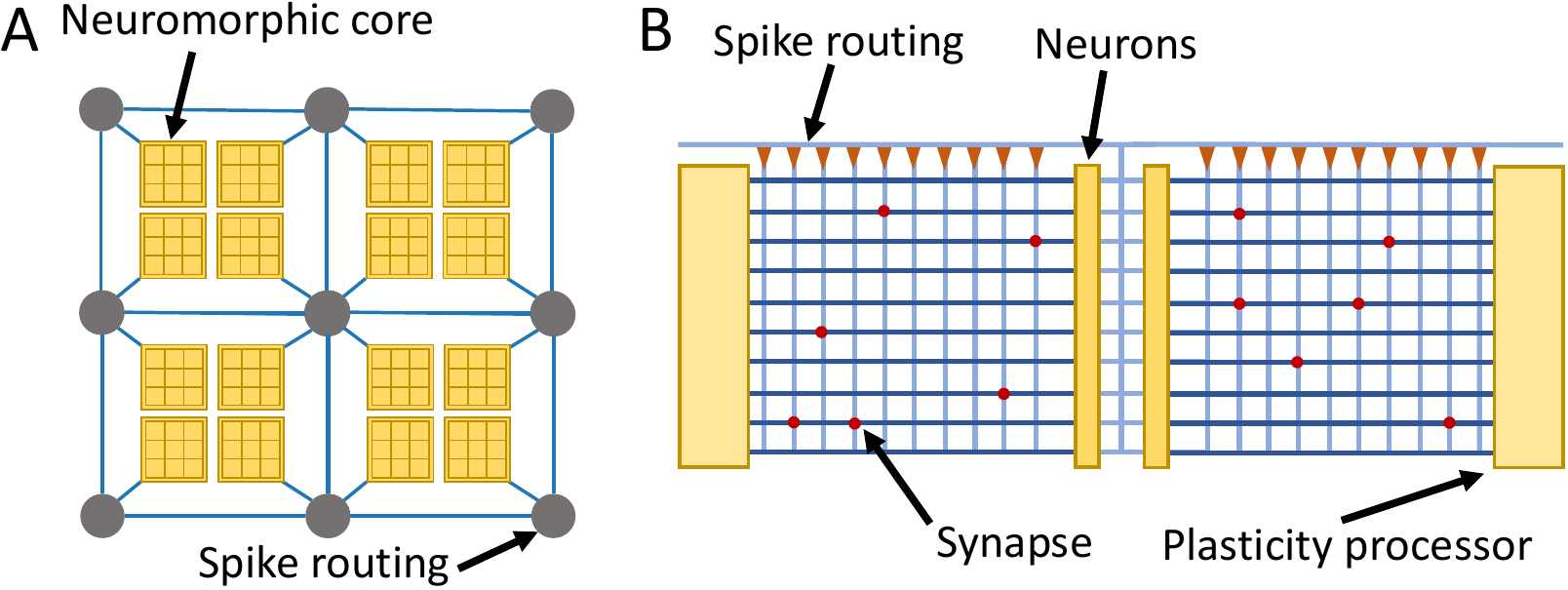}
    \caption{\textbf{(A)} Simplified schematic of the Loihi chip \cite{WikiChip_Loihi}. \textbf{(B)} Simplified schematic of the BrainScaleS-2 chip. Spikes emitted from the neurons enter the synaptic crossbar through synapse drivers (orange triangles) and, in essence, arrive at the target neuron when a connection is set (red dot). Each horizontal line in the crossbar is an input line for a neuron \cite{pehle2022brainscales}. \label{fig:NMHW}}
\end{figure}

First, Loihi's (\cref{fig:NMHW}A) main components are neuromorphic cores, which are processing units with custom circuitry and small amount of memory necessary to simulate a population of LIF neurons and their plasticity (i.e., learning rules).
Spikes are exchanged between neuromorphic cores using a routing grid.
In contrast, the BrainScaleS-2 chip (\cref{fig:NMHW}B) realizes physical AdEx neurons and synapses to emulate biology -- in other words, neurons and synapses are not \textit{simulated} but rather \textit{implemented} directly as analogue circuits.
Hence, no simulation time step exists and the system evolves continuously.
Although neurons and synapses are implemented using analogue circuits, spikes are transmitted digitally (also known as \lq\lq mixed-signal\rq\rq).
Through a synaptic crossbar, connections can be set flexibly and are not pre-wired.

The main potential of neuromorphic hardware lies in enabling the deployment of energy-efficient low-latency AI systems suitable for edge applications. First glimpses into this potential can already be obtained nowadays: in \cref{tab:neuromorphic}, we list a few neuromorphic platforms that provide both performance and energy benchmarks for MNIST classification, with most systems reaching an energy footprint on the order of $\mu$J per inference. We chose MNIST here as it is the most commonly used benchmark and allows for easy comparison across different (non-)neuromorphic platforms. However, it has been noted that SNNs perform best on data with temporal structure, meaning that MNIST is less suited as a benchmark for showing the advantages of SNNs as well as neuromorphic hardware \cite{zenke2021remarkable,davies2019benchmarks}. In fact, we argue that in order to better assess the potential of neuromorphic devices, standardized and application-specific benchmarks are required to fairly evaluate both algorithms and hardware platforms.

A recent feature of neuromorphic chips that is especially interesting for space applications is on-chip learning, found in chips like Loihi, BrainScaleS-2 and SPOON \cite{frenkel202028}. On-chip learning enables continuous learning directly integrated in the neuromorphic chip with a low energy footprint, suitable for the distributed design philosophy of neuromorphic devices. In the future, such `intelligent' neuromorphic devices could be especially useful for retraining or fine-tuning onboard models during deep space missions toautonomously adjust and adapt to previously unknown or unexpected circumstances.
\begin{table*}[ht]\centering
    \resizebox{\textwidth}{!}{\begin{tabular}{cccccc}
            \toprule
            Platform                                      & Type     & Tech  & Model    & MNIST     & Energy / Image                      \\
            \midrule
            Unnamed (Intel) \cite{chen20184096}           & digital  & 10nm  & LIF      & 97.70\%   & 1.7$\mu$J                           \\
            Intel Loihi \cite{davies2018loihi}            & digital  & 14nm  & LIF      & 96.40\%   & n.a.$^*$ \cite{lin2018programming}  \\
                                                          &          &       &          & 94.70\%   & 2.47mJ \cite{shrestha2021hardware}  \\
            IBM TrueNorth \cite{akopyan2015truenorth}     & digital  & 28nm  & LIF$^1$  & $92.70$\% & $0.268\mu$J                         \\
                                                          &          &       &          & 99.42\%   & 108$\mu$J                           \\
            Brainchip Akida \cite{vanarse2022application} & digital  & 28nm  & IF       & 99.20\%   & n.a.$^{*,2}$                        \\
            SpiNNaker-2 \cite{mayr2019spinnaker}          & digital  & 28nm  & --$^{3}$ & 96.60\%   & 23$\mu$J \cite{liu2018memory}       \\
            SPOON \cite{frenkel202028}                    & digital  & 28nm  & --$^4$   & 97.50\%   & 0.3$\mu$J$^5$                       \\
            BrainScales-2 \cite{pehle2022brainscales}     & analogue & 65nm  & AdEx     & $96.90$\% & 8.4$\mu$J \cite{goltz2021fast}      \\
            SpiNNaker \cite{furber2012overview}           & digital  & 130nm & --$^3$   & 95.00\%   & 3.3mJ \cite{stromatias2015scalable} \\
            \bottomrule
        \end{tabular}}
    \begin{flushleft}
        \tiny{$^*$ Not available.

            $^1$ \cite{akopyan2015truenorth} states that TrueNorth implements ``(...) a dual stochastic and deterministic neuron based on an augmented integrate-and-fire (IF) neuron model \cite{cassidy2013cognitive}''.

            $^2$ \url{https://doc.brainchipinc.com/zoo_performances.html}. Accessed: 2022-11-07.

            $^3$ SpiNNaker is an ARM-based processor platform and can therefore support arbitrary neuron models as long as a software implementation for SpiNNaker is available.

            $^4$ SPOON is an ``event-driven convolutional neural network (eCNN) for adaptive edge computing'' where ``TTFS encoding'' is used ``in the convolutional layers'' \cite{frenkel202028}.

            $^5$ Pre-silicon simulation result.}
    \end{flushleft}\vspace{-5mm}
    \caption{A list of neuromorphic platforms benchmarked on MNIST. As a baseline, \cite{goltz2021fast} provides values for a convolutional ANN implemented on a nVidia Tesla P100, reaching $99.2\%$ accuracy with an average energy / image of $852 \mu$J. It should be noted that all neuromorphic platforms here process images sequentially, while the GPU uses batching. For sequential processing, \cite{shrestha2021hardware} reports an accuracy of $98.90\%$ with an average energy / image of $37mJ$ (Intel i7 8700) and $16mJ$ (nVidia RTX 5000). Of the results shown above, only TrueNorth, Akida and SPOON implement a convolutional SNN. Table adapted from \cite{goltz2021fast}.}\label{tab:neuromorphic}
\end{table*}

\section{Application to onboard processing for Earth observation}\label{sec:nmhwappl}

As previously mentioned, the potential advantages of SNNs (e.g., in terms of energy efficiency) are promising for data processing, especially onboard miniaturized satellites. However, when applied to static data, as is the case in the classification of Earth observation images, it is not clear how SNNs compare to ANNs in terms of energy, latency and performance trade-offs \cite{kucik2021investigating}. For instance, some previous works \cite{bouvier2019spiking,han2017cross} showcase that when rate coding is used, the advantages of SNNs in terms of energy/performance decrease for classification datasets which contain complex features.
This makes satellite data a challenging task for SNNs because of the complexity of their features \cite{kucik2021investigating}.
For that reason, ESA's ACT, in collaboration with ESA's $\Phi$-lab, is currently benchmarking SNNs and ANNs for onboard scene classification. The latter was chosen as a target application due to the abundance of benchmark datasets such as EuroSAT \cite{helber2019eurosat,helber2018introducing} which was also the dataset chosen by the ACT for this project. EuroSAT is a 10-classes land-user and land-cover classification dataset consisting of images captured by the Sentinel-2 catellite. For the study, the RGB version of the images was used.
We are mostly investigating different information encoding solutions (including rate encoding and spike-time encoding) and their impact on the energy and accuracy that could be expected in performing onboard inference using an end-to-end neuromorphic approach. The results of past and ongoing efforts are presented in Sections \ref{subsubsec: RateCodingSNN4EO} and \ref{subsubsec: TimeCodingSNN4EO}, respectively.

\subsubsection{Rate-based SNNs for onboard Earth observation}
\label{subsubsec: RateCodingSNN4EO}
\begin{figure}[tb]
    \centering
    \includegraphics[width=0.9\linewidth]{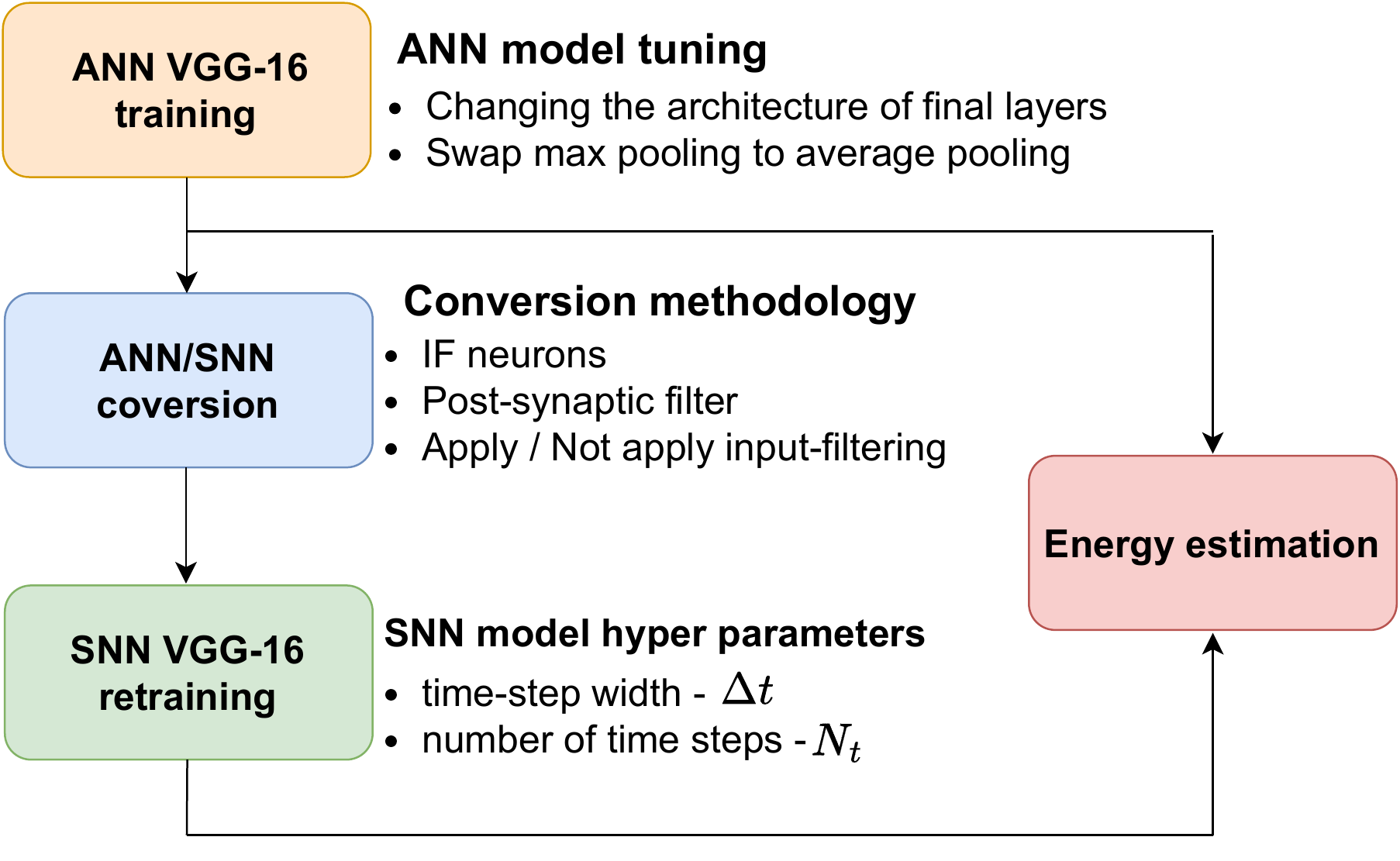}
    \caption{Rate-based SNN vs ANN performance/energy benchmark methodology used in our previous work \cite{kucik2021investigating}.}
    \label{fig: SNNRatedBased}
\end{figure}

Rate-based models were explored in a previous work by the ACT \cite{kucik2021investigating}, where the SNN model was trained by exploiting the approximate equivalence between spike rates of IF neurons and ReLU activations of artificial neurons, fundamentally using a weight conversion technique as discussed in \cref{sec:learningSNN} with limited loss in accuracy \cite{kucik2021investigating}. The methodology is summarized in Fig. \ref{fig: SNNRatedBased}.

As a first step, we started from a VGG16 ANN model \cite{simonyan2014very} pretrained on ILSVRC-2012 (ImageNet) \cite{deng2009imagenet}. We replaced the last three dense layers with a cascade of average pooling and dense layers. Furthermore, we replaced the max pooling with average pooling layers that do not require lateral connections and offer an easier implementation for spiking models \cite{kucik2021investigating}.
The VGG16 network was then trained on the EuroSAT RGB dataset.  
To convert the trained ANN model, we replaced the ReLu activation function with a cascade of IF neurons and post-synaptic filters, whose dynamic behaviour is described in \cite{kucik2021investigating}

\begin{equation}
    y(t) := \left(1 -e^{-\Delta t/\tau}\right) \cdot x(t) + e^{-\Delta t/\tau} \cdot x(t-1),
    \label{eq: PostSynapticLP}
\end{equation}
where $x(t)$ is the output spike train of a neuron at time $t$, $y(t)$ is the output of the postsynaptic filter, and $\Delta t$ is the timestep width.
Finally, the converted SNN model is retrained to optimize the performance. 
To estimate and compare the energy consumption of the SNN and ANN models, we exploited the methodology implemented in \textit{KerasSpiking}. 
In particular, more than performing an accurate estimation of the energy consumption, the methodology used aimed at allowing a relative comparison \cite{kucik2021investigating}. By considering a single spiking layer $L$, the used methodology assumes that the energy consumption during the inference is mostly due to two contributions: energy dissipated due to the synaptic activity $E_s$ and energy usage due to neuron updates $E_n$.
$E_s$ can be calculated as

\begin{equation}
    E_s = E_o\cdot \sum_N S_N \cdot f_{\text{in}} \cdot N_t \cdot {\Delta} t\,,
    \label{eq: Es}
\end{equation}
where $E_o$ is the energy for a single synaptic operation and depends on the hardware used, $S_{N}$ is the number of synapses for every neuron, $f_{\text{in}}$ is the average neuron spiking rate, $N_t$ is the number of timesteps in the simulation and ${\Delta}t$ is the timestep width.
For artificial models, $f_{\text{in}} = \frac{1}{{\Delta} t }$ and $N_t = 1$. 
Furthermore, $E_n$ can be calculated as

\begin{equation}
    E_n = E_u \cdot N_n \cdot N_t \,,
    \label{eq: E_n}
\end{equation}
where $E_u$ is the energy for a neuron update and depends on the hardware (similarly to $E_o$), and $N_n$ is the number of neurons in a layer. 
The total energy consumption for a model is then given by the sum of the contributions of the different layers. 
Since for the SNN models, the energy depends on $N_t$ and ${\Delta} t $, different SNN models were trained to explore various energy/accuracy trade-offs. To that end, we trained different models by setting the number of timesteps $N_t$ ranging from $1$ to $32$ and making ${\Delta} t $ a learnable parameter. The constant $\tau$ of every neuron post-synaptic filter was also trained to optimize the dynamic response of each neuron. 
In addition, since $E_s$ is proportional to $f_{\text{in}}$. Therefore, in order to reduce the average rate for each neuron, the input was processed with a Prewitt filter, whose dynamics is given by

\begin{equation}
    \mathbf{X}' := \max \left( c\sqrt{(\mathbf{G} \circledast \mathbf{X})^2 + (\mathbf{G}^\top \circledast \mathbf{X})^2}, \,  0.0078 \cdot \mathbb{1} \right),
    \label{eq:prewitt}
\end{equation}
where 
$$
    \mathbf{G}:= \begin{pmatrix}1 & 1 & 1\end{pmatrix}^\top \begin{pmatrix}1 & 0 & -1\end{pmatrix},
$$
and $\circledast$ represents the two-dimensional convolution operator, $\mathbb{1}$ is a unitary tensor having the same shape as the input $\mathbf{X}$, and $c $ is a normalization constant.
The effect of the Prewitt filter is to concentrate the input spikes mostly around the boundaries of different crops while keeping the input color information. Training was performed with and without applying the Prewitt filter to test its effects. 
Since the energy depends on the hardware used through the constants $E_o$ and $E_u$, to estimate the energy consumption we tested various models on different hardware, including on desktop CPU (Intel i7-4960X), one embedded processor (ARM Cortex-A), one GPU (nVidia GTX Titan Black) for ANN models and SpiNNaker \cite{furber2012overview}, SpiNNaker 2 \cite{mayr2019spinnaker}, and Loihi \cite{davies2018loihi} neuromorphic processors for both the ANN and SNN networks. 
The test results are summarized in Table \ref{tab: SNNResults}.

\begin{table}[tbp]
\begin{tabular}{|c|c|c|c|c|c|}
\hline
\textbf{Model} &
  \textbf{\begin{tabular}[c]{@{}c@{}}Accuracy \\ {[}\%{]}\end{tabular}} &
  \textbf{T} &
  \textbf{\begin{tabular}[c]{@{}c@{}}$\Delta t$ \\ {[}s{]}\end{tabular}} &
  \textbf{\begin{tabular}[c]{@{}c@{}}Energy \\ {[}J{]} \\ on GPU\end{tabular}} &
  \textbf{\begin{tabular}[c]{@{}c@{}}Energy\\  {[}J{]} \\ on Loihi\end{tabular}} \\ \hline
ANN                                                                     & 95.07 & 1 & -      & 0.06996 & 0.00636 \\ \hline
ANN + Prewitt                                                           & 90.19 & 1 & -      & 0.06996 & 0.00636 \\ \hline
\begin{tabular}[c]{@{}c@{}}SNN\\ (Best accuracy)\end{tabular}           & 85.11 & 4 & 0.0381 & -       & 0.00444 \\ \hline
\begin{tabular}[c]{@{}c@{}}SNN + Prewitt\\ (Lowest energy)\end{tabular}   & 87.89 & 4 & 0.0403 & -       & 0.00205 \\ \hline
\begin{tabular}[c]{@{}c@{}}SNN + Prewitt\\ (Highest accuracy)\end{tabular} & 85.07 & 1 & 0.0813 & -       & 0.00476 \\ \hline
\end{tabular}
\caption{Summary of the results shown in Table 2 of \cite{kucik2021investigating}.}
\label{tab: SNNResults}
\end{table}

The maximum accuracy of the SNN model was $85.11\%$ and $87.89\%$ with and without Prewitt filtering, respectively \cite{kucik2021investigating}. The ANN models reached $95.07\%$ and $90.19\%$ with and without Prewitt filter, respectively. 
In addition, the input filter had a strong effect on the tradeoff between energy consumption and accuracy for SNN models, whereby the energy consumption dropped by half while maintaining similar accuracy or the accuracy increased by over $2\%$ with a similar energy consumption. 
The drop in accuracy of the SNN + Prewitt (highest accuracy) and SNN + Prewitt (lowest energy) was $1.18\%$ and $10\%$, respectively, which was significant. However, when compared to the GPU implementation of the ANN, inference with SNN + Prewitt (highest accuracy) and SNN + Prewitt (lowest energy) on Loihi required $14.70$ and $34.13$ times less energy, respectively. 
Since this improvement was partially because of the higher efficiency of Loihi compared to the GPU, we also tested the inference performance of the ANN model on Loihi to eliminate the effects of the hardware. Even in this case, the energy consumption was $3.1$ and $1.3$ times lower compared to the SNN + Prewitt (lowest energy) and SNN + Prewitt (highest accuracy) models. 
However, to obtain more realistic estimates, future work will aim at more hardware-aware proxies that also take into account other effects, such as on-chip saturation effects and power-consumption due to off-chip accesses, which might affect the estimates presented here. In addition, the results seem to confirm trends described in \cite{bouvier2019spiking,han2017cross}, according to which longer simulation leads to better accuracy at the expense of energy consumption when rate coding is used. Because of that, as described in Section \ref{subsubsec: TimeCodingSNN4EO}, different information encoding solutions shall be explored.

\subsection{Spike-time-based SNNs for onboard Earth observation}
\label{subsubsec: TimeCodingSNN4EO}

Additional encoding schemes, such as temporal encoding, and surrogate gradient training approaches are currently being explored that could yield better trade-offs for Earth observation. 
By encoding information in spike times instead of rates, information can be processed with a much lower number of spikes, leading to higher sparsity in the network activity and consequently lower energy consumption.
We are focusing on two approaches: (i) training SNNs that solely use TTFS encoding, either trained with exact gradients or surrogate gradients, and (ii) training SNNs end-to-end using surrogate gradients without imposing any specific encoding scheme.
Our evaluation of models is two-fold: performance and network metrics (for instance, the average number of spikes per inference) are investigated in simulation using SNN frameworks such as Norse.
In addition, we are mapping parts of these models to the neuromorphic chips Loihi and BrainScaleS to obtain an estimate for the energy efficiency of a fully neuromorphic implementation.
As this is still an ongoing study, only a few preliminary results will be discussed here.

Performance-wise, we found that convolutional architectures trained end-to-end show the highest accuracy.
We implemented a spike-based convolutional neural network (CNN) with 4 layers (3 convolutional, 1 dense layer) that takes the pixel values of images (rescaled to be between $0$ and $1$) as input currents.
In this architecture, the output layer consists of LIF neurons without a threshold mechanism, i.e., without spiking.
All other layers consist of LIF neurons.
A classification result is given by the index of the neuron with the maximum membrane potential value during the whole simulation time -- in simpler terms, the label neuron that obtained most votes from the last hidden layer provides the classification outcome.
Trained end-to-end, this model achieves competitive results ($\approx 91-92\%$ test accuracy on EuroSAT RGB) with low spike activity (on average $\approx 2 - 3$ spikes per neuron during inference).

Moreover, we are investigating SNNs based on TTFS encoding for all layers (input, hidden and output).
Thus, in the label layer, the classification outcome is given by the label neuron that spikes first.
We are currently exploring both models that can be trained using exact gradients and ones that can be trained using surrogate gradients.
Preliminary results show that these models reach slightly lower accuracy, although they also require fewer spikes per inference than models without pre-imposed encoding.
In general, these initial results already hint at a confirmation that direct training of SNNs using surrogate gradients leads to better results than (rate-based) weight conversion for Earth observation image processing tasks.

\section{Neuromorphic sensing}
\label{subsec: eventSensing}

In addition to \textit{processing}, neuromorphic computing has also entered the domain of \textit{perception}, most notably in the area of computer vision. Biological systems provide a rich source of inspiration due to their energy efficiency, strong data compression and feature extraction properties as well as remarkable adaptability. 
A widely used example of biological computations is the visual system in flying insects, which has been deeply researched in the last decades (e.g. \cite{franceschini1992insect, song2013digital}). Research on the neuronal systems in insects allows for single-neuron analysis and thus provides very detailed insights on the internal working mechanisms of insect vision. 
The primary source of motion information in insects is the so-called EMD - Elementary Motion Detector (e.g. \cite{frye2015elementary}). EMDs are a minimalistic neural architecture able to extract important motion information from only two photoreceptors. Preliminary studies performed at the European Space Agency on their applications to spacecraft landing \cite{valette2010biomimetic, izzo2011constant, Medici2010Neuromorphic, Valette2009Neuromorphic} have suggested the possibility to use EMDs as the only needed sensor to successfully land spacecraft in specific scenarios.
A second example of highly efficient and adaptive biological sensor is the mammalian retina, which is composed of several layers of neurons specialised in processing visual information. The mammalian visual system is important to understand the biological inspiration behind the concept of event-based vision, which we later focus on, thus the following section provides a brief overview of its structure and function.

\subsection{Structural and functional organisation of the retina}\label{sec:retina.structure_and_function}
The structural and functional organisation of the mammalian retina has been explored in great detail over the past few decades. The elaborate 1965 study by Hubel and Wiesel \cite{HubelWiesel_1965_RECEPTIVEFIELDSFUNCTIONAL} was the first to present a collective summary of many fundamental properties of the retina, such as the hierarchy of retinal neurons, their receptive fields (RFs) and orientation selectivity, and so forth. The mammalian retina consists of several layers containing nerve cells and receptors that perform highly specialised functions \cite{Grigoryan_2022_SelfOrganizationRetinaEye}. The following is a brief overview of the main types of cells found in the mammalian retina, together with their main distinctive features.

\paragraph{Receptors}\label{par:retina.receptors}
The outer nuclear layer (ONL) contains the actual photosensitive receptors: rods, which are responsible for scotopic vision (low-light conditions), and cones, which are responsible for photopic vision (in well-lit conditions). Furthermore, rods function in low-light conditions but do not distinguish colour, whereas cones are active in display colour specialisation (e.g., responding to light in the red, green or blue part of the visible spectrum). Receptors are excitatory cells.

\paragraph{Horizontal cells}\label{par:retina.horizontal_cells}
The signal produced by receptors in the ONL is modulated by horizontal cells, which provide local inhibitory signal informed by the immediate neighbourhood of the receptor. Horizontal cells are responsible for local brightness adaptation (especially in the foveal region) as well as the formation of ON/OFF RFs in ganglion cells \cite{Masland_2011_CellPopulationsRetina}).

\paragraph{Bipolar cells}\label{par:retina.bipolar_cells}
The raw visual input normalised through feedback from horizontal cells is then fed into bipolar cells, which effectively split the visual channel into two pathways (ON and OFF) \cite{Masland_2011_CellPopulationsRetina}. This means that ON bipolar cells respond to an increase in brightness, whereas OFF bipolar cells respond to a \textit{decrease} in brightness. The separate pathways are preserved all the way to the visual cortex and have been shown to be asymmetric in terms of their behaviour -- for instance, the OFF pathway reacts faster and sends stronger signals to the cortex compared to the ON pathway \cite{MazadeJinPonsEtAl_2019_FunctionalSpecializationCortical}. In addition, while horizontal cells provide the retina with spatial brightness adaptation capabilities, bipolar cells are responsible for temporal adaptation and filtering \cite{EulerHaverkampSchubertEtAl_2014_RetinalBipolarCells,GollischMeister_2010_EyeSmarterScientists}.

\paragraph{Amacrine cells}\label{par:retina.amacrine_cells}
The excitatory signals produced by bipolar cells are combined with inhibitory feedback from amacrine cells in many different configurations. There are more than 30 types of amacrine cells, whose primary function is to combine the signals arriving from bipolar cells in many different configurations that define, modulate and control the RFs of ganglion cells via inhibitory synapses \cite{GollischMeister_2010_EyeSmarterScientists} (however, there is a possibility that they also provide excitatory signals via gap junctions \cite{Masland_2012_TasksAmacrineCells}).

\paragraph{Ganglion cells}\label{par:retina.ganglion_cells}
Amacrine and bipolar cells form different wiring patterns providing input to the ganglion cell layer (GCL). Different ganglion cells have different RFs, which determine their \textit{selectivity} to certain patterns (both static and dynamic, such as orientation, motion or combinations of both) in the visual field \cite{Masland_2011_CellPopulationsRetina}. For instance, some ganglion cells respond to edges, others are highly sensitive to motion relative to the background in the visual field, and yet others respond to approaching motion while remaining silent in response to lateral motion \cite{GollischMeister_2010_EyeSmarterScientists}. While ganglion cells are officially considered to be the first neurons in the retina that produce spikes \cite{ChichilniskyKalmar_2003_TemporalResolutionEnsemblea} (the other cells have graded responses), there are indications that at least bipolar cells also produce a limited number of spikes \cite{BadenEspostiNikolaevEtAl_2011_SpikesRetinalBipolarb} that are phase-locked to visual stimuli.

\par
Counterintuitively, the ONL is the \textit{innermost} layer of the retina, which means that light has to travel through all the other layers before it reaches the receptors. This and other structural properties of cellular layers in the retina have several functional implications.

\begin{itemize}
    \item{The GCL contains a lot of cell bodies as well as axons which relay spikes via the optic nerve to the visual cortex. Since the GCL is on top of the receptor layer, the part of the retina where ganglion cell axons interface with optic nerve is devoid of receptors, giving rise to the infamous \textit{blind spot}.}
    \item{A small area of the retina known as the \textit{fovea}, which is located at the centre of the retina directly behind the lens, contains the highest concentration of receptors. To maximise the resolution in the fovea, the axons of ganglion cells in the fovea are highly stretched out, forming a dense circular bunch of axon bodies around the fovea, where the resolution starts to drop sharply.}
    \item{At least in humans (and many other mammals), most of the colour-sensitive receptors are in the foveal region. Away from the fovea, vision is dominated by rods, which are much more sensitive (down to a single photon \cite{TinsleyMolodtsovPrevedelEtAl_2016_DirectDetectionSingle}) but not colour-specific. This is easily observed in low-light conditions, where our visual perception shifts towards detect only silhouettes).}
    \item{The RFs of ganglion cells responsible for encoding signals from the fovea are connected to as few as a single receptor, while the RF size increases further away from the fovea. This eccentricity-dependent RF size has been found to have a number of interesting consequences for scene recognition, including cortical magnification, scale invariance and object recognition \cite{HanRoigGeigerEtAl_2020_ScaleTranslationinvarianceNovel,PramodKattiArun_2022_HumanPeripheralBlur}.}
\end{itemize}

\subsection{Adaptation and homeostasis in the retina}\label{sec:retina.adaptation}

Adaptation is a hallmark of most types of neurons and circuits found in the retina, a lot of research effort has been dedicated to unveiling the underlying principles of visual adaptation \cite{BarlowFoldiak_1989_AdaptationDecorrelationCortex,ShapleyEnroth-Cugell_1984_VisualAdaptationRetinal}. A number of different homeostatic processes have been observed in practically all layers in the retina. For instance, cones have a very large dynamic range and are also extremely sensitive to small changes in contrast, both in terms of response time and contrast range. In comparison, rods can detect very low levels of illumination (as low as individual quanta of light \cite{BurnsLamb_2004_VisualTransductionRod}), with the trade-off that they adapt very slowly and become saturated by large sudden changes in light intensity \cite{BurnsLamb_2004_VisualTransductionRod}. ON-type (OFF-type) bipolar cells have a light center/dark surround (dark center/light surround) and respond strongly to positive (negative) contrast. The ON/OFF differentiation is maintained in retinal ganglion cells (RGCs), which receive input from one of the two types of BCs. The RFs of both ON and OFF RGCs independently cover almost the entire retina \cite{RatliffBorghuisKaoEtAl_2010_RetinaStructuredProcess}. The retina uses adaptation mechanisms in BCs and RGCs to adapt to both spatial and temporal variations in illumination \cite{FreemanGranaPassaglia_2010_RetinalGanglionCell,Rieke_2001_TemporalContrastAdaptation}. Both BCs and RGCs homeostatically regulate the intensity of their output depending on the magnitude of the stimulus, allowing the retina to adapt to changes in luminance that span $ 9 $ to $ 12 $ orders of magnitude over a 24-hour period \cite{RiderHenningStockman_2019_LightAdaptationControls,PearsonKerschensteiner_2015_AmbientIlluminationSwitches}.

Importantly, there is overwhelming evidence that RGCs adapt their activity to both the mean and the variance of the input \cite{FreemanGranaPassaglia_2010_RetinalGanglionCell,RiekeRudd_2009_ChallengesNaturalImages,RiderHenningStockman_2019_LightAdaptationControls}. In addition to temporal adaptation, both ON and OFF BCs also maintain a fast `push-pull' mechanism mechanism that balances the activation of the center relative to the surround for the purpose of locally enhancing the contrast within the RF \cite{MeisterBerry_1999_NeuralCodeRetina,VanWykWassleTaylor_2009_ReceptiveFieldProperties}. There is even some evidence that RGCs also employ a gain control mechanism to change the effective \textit{size} of the RF based on the illumination \cite{BarlowFitzhughKuffler_1957_ChangeOrganizationReceptive}.

Finally, the eye is not a static sensor -- it is constantly performing a combination of motions at various spatial and temporal scales -- ocular tremors, drift, microsaccades and saccades \cite{Martinez-CondeMacknik_2017_UnchangingVisionsEffects}. These different types of movements serve different purposes: for instance, small random movements (such as drift and microsaccades), which are outside voluntary control and occur even while fixating on a target, prevent the phenomenon of visual fading (also known as retinal fatigue), in which the rapid adaptation of the retinal ganglion cells to a fixed visual scene leads to the complete elimination of neural response (the eye effectively becomes blind to anything that is not moving) \cite{Martinez-CondeMacknik_2017_UnchangingVisionsEffects,AlexanderMartinez-Conde_2019_FixationalEyeMovements}.
The concept of a vision sensor modelled around the retina emerged more than three decades ago by building upon detailed research on the neuroscience of the mammalian retina. The following section presents a brief overview of the major milestones since the conception of the earliest retinomorphic circuits to present-day commercial event-based cameras.

\subsection{Retinomorphic models}\label{sec:retinomorphic}
Neuromorphic systems that specifically deal with modelling the mammalian visual system (mostly the retina) are known as `retinomorphic' systems \cite{Boahen_1996_RetinomorphicVisionSystems}. While some the operating principles of the circuits differ substantially from that of biological retinal cells, they are in many respects \textit{functionally} identical (or at least very similar) to biological retinal cells.

To optimize the power consumption and response speed of sensors, most recent research on event-based vision is heavily oriented towards hardware solutions, which are difficult to customise for the purpose of implementing novel computer vision algorithms and processing paradigms. In addition, existing software libraries are geared towards processing and emulating the existing hardware rather than providing a generic base for implementing novel algorithms.

The ACT is thus pursuing research on retinomorphic models to explore the question of whether sophisticated visual processing can be performed \textit{on the sensor}, rather than being \textit{learned} by a downstream algorithm. A prime example can be given with specialised ganglion cells which are sensitive to certain types of motion in their RF, to the exclusion of all other types of motion. For instance, there are ganglion cells that are sensitive to approaching motion while being insensitive to approaching or receding motion \cite{GollischMeister_2010_EyeSmarterScientists,ApplebyManookin_2020_SelectivityApproachingMotiona}, which can be useful for applications such as spacecraft landing, where the landscape is perceived as approaching from the perspective of the camera.

To maximise the efficiency and speed of onboard processing, it would be beneficial to design sensors capable of obtaining insight about motion and features in the visual field, such as edges, depth or optical flow. We know that biological ganglion cells with different RFs are sensitive only to very specific features and types of motion in the visual field (such as approaching motion or egomotion) while largely remaining silent in the presence of other types of motion (such as receding or lateral motion). It would therefore make sense to be able to model the function of these cells in a convenient and reproducible way that could ultimately inform the design of novel dedicated hardware sensors.

Therefore, the ACT is developing a simulation framework that can model as many features of the mammalian retina as possible in an efficient manner in order to facilitate rapid prototyping of retinomorphic architectures. In particular, our research and development efforts have been geared towards implementing support for all of the following features:

\begin{itemize}
    \item{All major types of cells in the retina (cf. \ref{sec:retina.structure_and_function});}
    \item{Spatial and temporal adaptation and homeostasis (cf. \ref{sec:retina.adaptation});}
    \item{Complex RFs (e.g., centre/surround);}
    \item{Foveation and eccentricity-dependent RFs;}
    \item{Sparsification of cells towards the edges of the retina (i.e., cell distribution becomes increasingly sparse towards the edges of the retina, in proportion to increasing RF size).}
    \item{Saccadic movements (microsaccades, drift, etc.);}
\end{itemize}

As outlined above in Section \ref{sec:retina.structure_and_function}, the retina consists of several layers, whereby cells in each layer (except the photoreceptor layer) receive input from a certain local neighbourhood of cells cells in the preceding layer(s). This mechanism is behind the inspiration for CNNs, which have occupied a central role in computer vision domain of AI research \cite{LeCunBoserDenkerEtAl_1989_BackpropagationAppliedHandwritten,KrizhevskySutskeverHinton_2012_ImageNetClassificationDeep}. The successful application of CNNs on a wide range of tasks, combined with the availability of several software packages that offer highly optimised CNN implementations out of the box, has largely dampened any incentive to explore alternative architectures or mechanisms for implementing convolution. However, incorporating all of the above features into a coherent framework is not straightforward, specifically with respect to saccadic movements and eccentricity-dependent RFs. This requires rethinking how convolution is implemented.

\subsubsection{Sparse convolution}\label{sec:sparse_convolution}

Convolution is a costly operation to implement, and state-of-the-art CNNs rely on a number of tricks to speed up the process. $N$-dimensional convolution is usually implemented using of a number of convolutional kernels (represented as a single $(N + 2)$-dimensional tensor, where $N$ is the number of dimensions of the input and two extra dimensions representing the number of kernels and the batch size). Therefore, without loss of generality, we focus on $ 2D $ convolution, unless explicitly stated otherwise.

The generic expression for convolving a batch of $ N $ $ 2D $ inputs (each with dimensions $ H $ and $ W $) using $ C_{in} $ convolutional kernels can be given as follows:

\begin{align}
    Conv2d(N, C_{out}, H, W) = \sum_{i=0}^{C_{in}}{W_{(C_{out},i)} \star Input(N, i)}
    \label{eq:generic_convolution}
\end{align}

The dimensions $ H_{out} $ and $ W_{out} $ of the output channels are computed as

\begin{align}
    H_{out} = \left\lfloor \frac{H_{in} + 2p_{H} - d_{H}(k_{H} - 1) -1}{s_{H}} + 1 \right\rfloor \label{conv.height} \\
    W_{out} = \left\lfloor \frac{W_{in} + 2p_{W} - d_{W}(k_{W} - 1) -1}{s_{W}} + 1 \right\rfloor \label{conv.width},
\end{align}

Here, $p$, $d$, $k$ and $s$ (with indices $H$ and $W$) represent the padding, dilation, kernel size and stride of the convolution operation (in the direction of $H$ and $W$, respectively). These four parameters impose constraints on each other depending on the input since $H_{out}$ and $W_{out}$ in \cref{conv.height,conv.width} must be integer. For instance, convolving an image of size $32 \times 32 $ with a kernel of size $5 \times 5$, a stride of $2$ and dilation and padding of $0$ means that the convolution would not be symmetrical at the edges of the image since ensuring that $H_{out}$ and $W_{out}$ are integer enforces the convolution operation to stop short of covering the last column and row of the image. Adding a padding of $2$ around the image mitigates this problem and ensures that the convolution is symmetric. We refer the reader to the relevant literature for an in-depth overview of the meaning of these parameters \cite{GoodfellowBengioCourville_2016_DeepLearninga}.

\begin{figure}
    \begin{subfigure}[pbh]{0.48\textwidth}
        \centering
        \includegraphics[width=\linewidth]{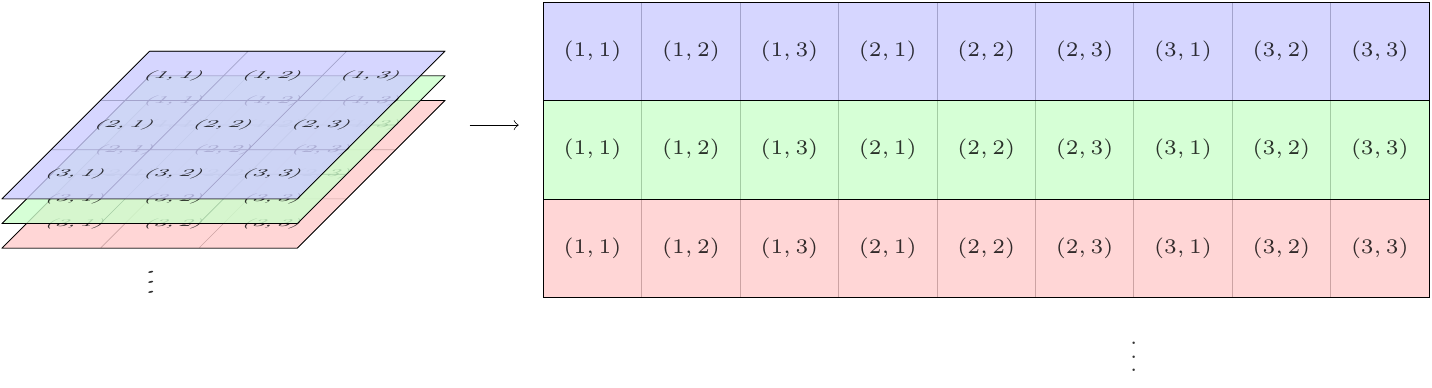}
        \caption{Kernel unfolding}\label{fig:kernelstretch}
    \end{subfigure}
    \hfill
    \begin{subfigure}[pbh]{0.48\textwidth}
        \centering
        \includegraphics[width=\linewidth]{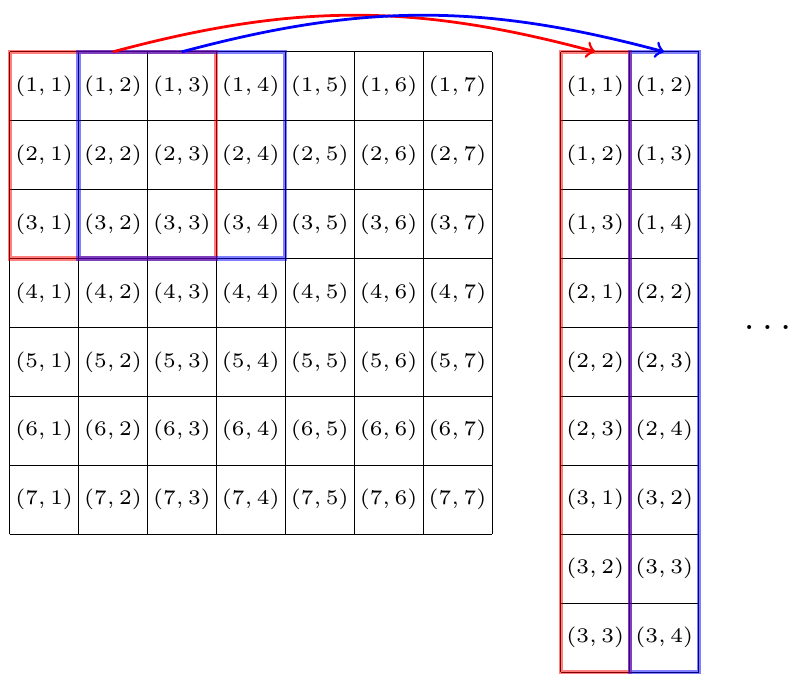}
        \caption{Input unfolding}\label{fig:colstretch}
    \end{subfigure}
    \begin{subfigure}[pbh]{\textwidth}
        \centering
        \includegraphics[width=0.9\linewidth]{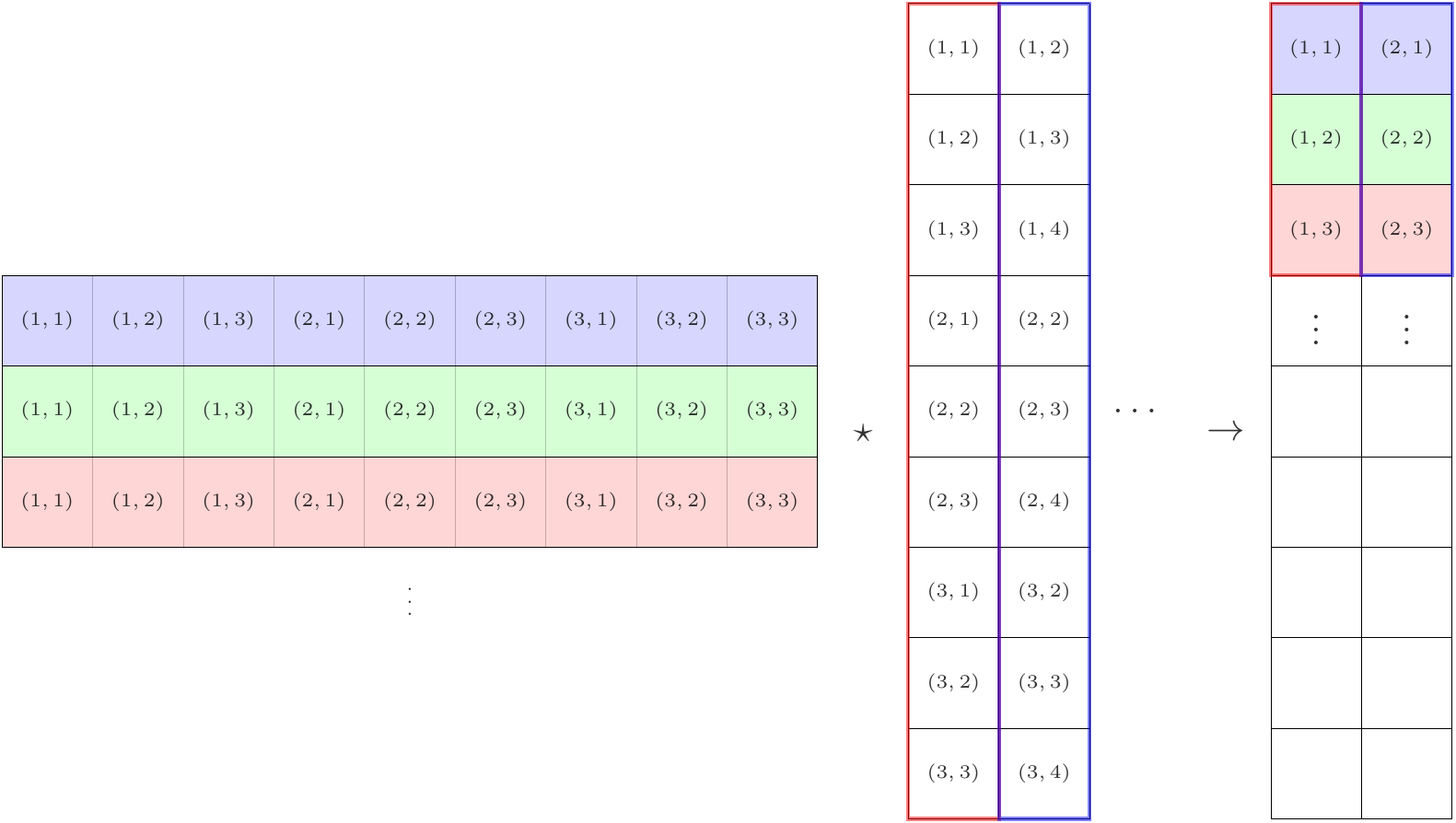}
        \caption{Convolution operation.}\label{fig:convolution}
    \end{subfigure}
    \caption{Default Im2col operation.}\label{fig:im2col}
\end{figure}

In practical implementations, convolution relies on a crucial but often understated preprocessing step. This operation, known as `image-to-column' (or $ im2col $ for short) \cite{ChellapillaPuriSimard_2006_HighPerformanceConvolutional} involves stretching out the convolutional kernels and arranging them into rows of a matrix, and stretching \textit{all} kernel-sized subsections of the input into one-dimensional column vectors and arranging them as columns of another matrix (hence the term $im2col$, since $2D$ convolutions are most commonly applied to images) (Fig. \ref{fig:im2col}). In this way, the convolution operation is reduced to a single dense matrix-matrix multiplication.

It is immediately clear that while it improves the efficiency of convolution, this approach introduces a number of cascading limitations that impede the effort to implement the desired features outlined in \ref{sec:retinomorphic}:

\begin{itemize}
    \item{All kernels must be the same size, which precludes foveation.}
    \item{An inability to convolve the input with kernels of different size limits the type of RFs that can be implemented; specifically, centre/surround RFs are difficult to implement with a fixed kernel size, especially in combination with foveation.}
    \item{Sparsification is impossible to model since the operation is by definition \textbf{dense}, so it applies the kernel to all segments of the input as defined in \cref{eq:generic_convolution}.}
    \item{It is not trivial to implement saccadic movements as this would entail performing the $ im2col $ preprocessing step for each movement. Furthermore, as an attention shifting mechanism, saccadic movements are much more useful in combination with foveation.}
    \item{Without saccades, adaptation (specifically temporal adaptation) as observed in the retina quickly erases features in stationary parts of the image.}
\end{itemize}

\begin{figure}
    \begin{subfigure}[pbh]{\textwidth}
        \centering
        \includegraphics[width=0.9\linewidth]{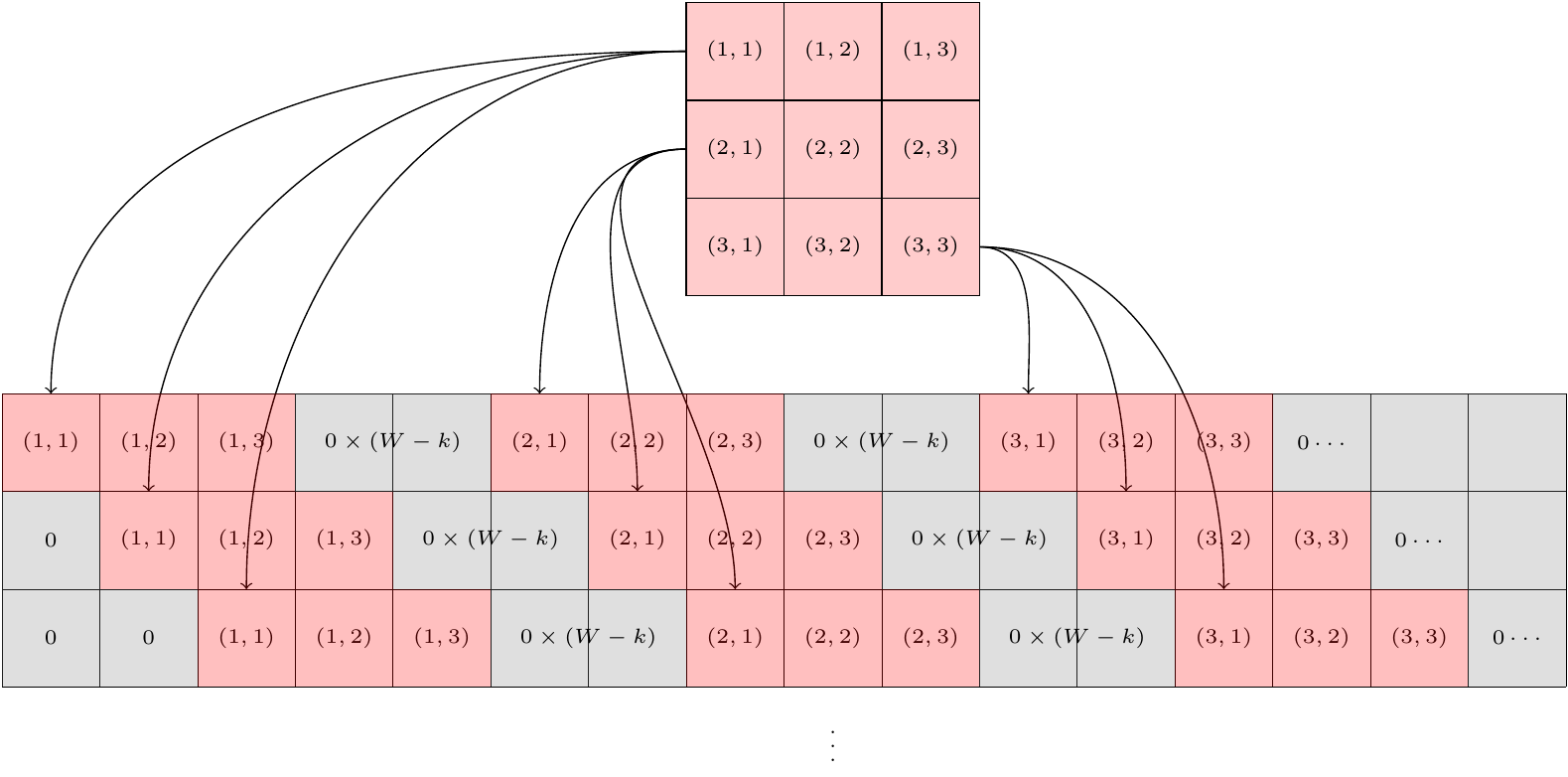}
    \end{subfigure}
    \caption{Unfolding a kernel into a sparse matrix. The space redundancy introduced in this operation is offset by the elimination of the $ im2col $ operation.}\label{fig:k2row}
\end{figure}

To address these limitations, we have developed a straightforward alternative to the default convolution operation. Rather than unfolding the input, this alternative convolutional method relies on a \textit{kernel} unfolding step (tentatively named $kernel2row$, or $k2row$ for short) that consists of unfolding the kernels, splitting them into continuous chunks and arranging those chunks into a \textit{sparse} matrix, while the input is unfolded into a single large column vector. The kernel chunks are arranged in such a way that the result of multiplying the sparse kernel matrix and the stretched input vector is identical to that obtained with the above dense convolution method. The $k2row$ operation is outlined in Fig. \ref{fig:k2row}.

The most notable advantage of performing convolution in this way is that the kernel size can be decoupled from the implementation constraints, allowing for kernels of different sizes to be used within the same convolution matrix.

Most of the desired features for simulating a retinal model as outlined in \ref{sec:retinomorphic} can be implemented trivially with this sparse convolution method. For instance, foveation can be implemented by using smaller kernels towards the centre of the image than towards the edges. Similarly, a separate kernel `instance' can be used to convolve each pixel of a $2D$ image, which allows us to implement complex RFs (such as ones relying on the centre-surround mechanism) as a difference of Gaussians (`Mexican hat' filter) represented by the difference of two convolutions with kernels of different size. Finally, saccadic movements can be implemented in combination with foveation as an attention shifting mechanism by simply zero-padding the input image in a way that offsets the centre of the resulting padded image by the desired amount equal to the distance travelled by the `fovea' during the saccadic movement.

To demonstrate the flexibility of this method, we implemented a pipeline emulating all retinal layers (see \S\ref{sec:retina.structure_and_function}). In this model, the input to excitatory layers (receptors, bipolar cells and ganglion cells) is modulated by feedback from inhibitory layers (horizontal and amacrine cells). Cells in the excitatory layers have eccentricity-dependent RFs (smaller towards the centre of the image and gradually growing in size towards the edges). A local spatial normalisation map is implemented using the horizontal layer, whereas the bipolar layer implements a temporal filter as well as the split into separate ON and OFF pathways. The temporal filter was implemented as an exponential running mean which was updated iteratively as follows at every frame:

\begin{align}
    \mu_{t+1} = \alpha x + (1-\alpha) \mu_{t},
\end{align}

where $\mu_{t}$ is the mean at time step $t$, $\alpha$ is the `forgetting rate' that determines how much of the preceding input is remembered, and $x$ is the input frame. The mean is computed in this way for each pixel.

The ganglion layer preserves the separation of these two paths and implements two types of ganglion cells with ON-centre / OFF-surround and OFF-centre / ON-surround RFs, respectively. For the sake of brevity, we have omitted the demonstration of saccadic movements, which can be used for inducing events from an otherwise static input. The parameters for this model are given in Table \ref{table:retinomorphic_model_parameters}. We use part of the Perseverance landing sequence as recorded with the rover's onboard camera minutes before touchdown \cite{perseverance_landing}.

\noindent
\begin{table}[hpt]
    \caption{Parameters for the retinomorphic model in \ref{sec:sparse_convolution}\label{table:retinomorphic_model_parameters}}
    \begin{threeparttable}
        \renewcommand{\tabularxcolumn}[1]{m{#1}}
        \renewcommand{\arraystretch}{1.4}
        \centering
        \begin{tabularx}{\textwidth}{l|X|r}
            \hline
            \hline
            Cell type & Parameter & Value  \\ 
            \hline
            Receptors & Input type & Grayscale \\
            \hline
            \multirow{4}{*}{\parbox{3cm}{Horizontal, bipolar, amacrine, ganglion}} & RF size distribution & Gaussian\tnote{1} \\
                                              & RF size distribution: mean  & $(H/2, W/2)$\tnote{2} \\
                                              & RF size distribution: SD    & $ (H/3, W/3)$ \\
                                              & RF type  & Proportional\tnote{3} \\
            \hline
            \multirow{2}{*}{\parbox{3cm}{Horizontal, bipolar, amacrine}} & Min. / max. RF size & $ 1 \times 1 $ / $ 4 \times 4$ \\
            & & \\
            \hline
            \multirow{2}{*}{\parbox{3cm}{Ganglion}} & Min. / max. RF size (centre) & $ 1 \times 1 $ / $ 4 \times 4$ \\
                                              & Min. / max. RF size (surround) & $ 4 \times 4$ / $9 \times 9$ \\
            \hline
            Bipolar cells  & $\alpha$\tnote{4} & $ 0.95$ \\
            \hline
            Ganglion cells  & Threshold\tnote{5} & 15 \\
            \hline
        \end{tabularx}
        \begin{tablenotes}
            \item[1]{For all cell types except horizontal cells, the Gaussian was \textit{inverted}, meaning that the RF size increased with distance from the centre. For horizontal cells, the RF size was \textit{larger} at the centre and decreased towards the edges. This is consistent with the fact that horizontal cells are concentrated around the foveal region \cite{Masland_2011_CellPopulationsRetina}, where they facilitate the mechanism of adaptation to the local brightness level. A larger RF size for horizontal cells in the foveal region leads to a sharper image in the fovea, consistent with observations from biology.}
            \item[2] {In this table, $H$ and $W$ denote the height and width of the input image, respectively.}
            \item[3] {Proportional here means that the value of each element of the kernel is the inverse of the kernel size (e.g., $1/9$ for a $3 \times 3$ kernel).}
            \item[4] {$\alpha$ is the `forgetting rate' of the exponential running average used as a temporal filter in the bipolar layer.}
            \item[5] {The threshold indicates the difference between the intensity at the centre vs. the surround of the receptive field. An event is produced if this threshold is crossed.}
        \end{tablenotes}
    \end{threeparttable}
\end{table}

\begin{figure}[htp]
    \centering
    \includegraphics[width=\textwidth]{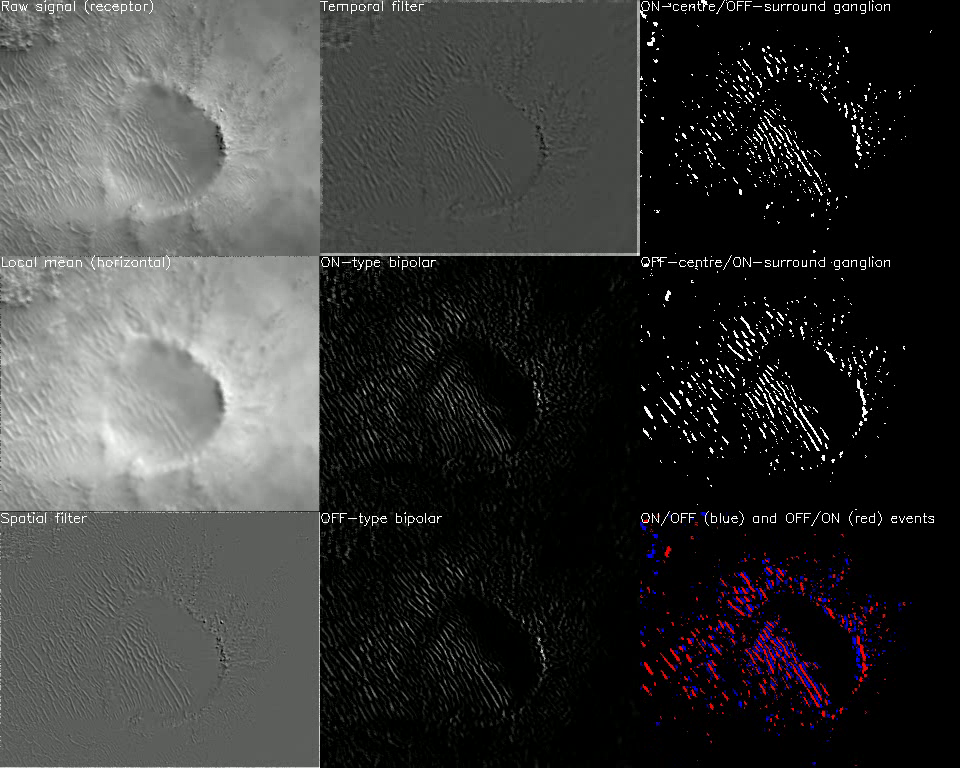}
    \caption{A snapshot of the state of each layer and the combination of ON/OFF and OFF/ON events. (a) Raw input (converted to grayscale). (b) Simulated horizontal cell layer computing the local mean illumination. (c) The raw signal with the mean illumination from (b) subtracted. (d) Temporal filter (implemented as a running average of the normalised signal in (c). (e) ON and (f) OFF bipolar cells computing positive and negative deviation from the temporal mean, respectively (i.e., parts that are brighter and darker than the mean). Ganglion cells with (g) ON/OFF and (h) OFF/ON centre/surround RFs. In ON/OFF ganglion cells, the centre is driven by ON bipolar cells and the surround by OFF bipolar cells, and vice versa for OFF/ON ganglion cells. A ganglion cell produces a binary spike (event) when the difference between the centre and surround crosses a predefined threshold (a model parameter). Finally, (i) shows the combined activity of ON/OFF (blue) and OFF/ON (red) ganglion cells.}\label{fig:retinal_model_all_layers}
\end{figure}


Several interesting observations can be made about the results in Fig. \ref{fig:retinal_model_all_layers}. First, despite the simplicity of the simulated retina, it can clearly perform certain types of preprocessing, such as local brightness normalisation at the horizontal layer, edge extraction (e.g., at the bipolar layer) and event generation at the ganglion layer. It is also noteworthy that ON/OFF and OFF/ON ganglion cells highlight the two sides (brighter and darker) of edges in the visual scene. 
It is noteworthy that the RF organisation in this case successfully suppresses noise in the output of the ganglion layer. This demonstration of effects observable in biological retinas is encouraging as we are working towards implementing more sophisticated processing of the incoming visual information, such as optical flow reconstruction and detection of approaching motion for applications such as obstacle avoidance and landing.

The retinal simulation code is part of a library named $pyrception$, which in turn is part of the ACT open-source ecosystem \footnote{\url{https://gitlab.com/cantordust/pyrception}}. The ultimate goal of the $pyrception$ library is to provide convenient interfaces for converting input from multiple sensory modalities (such as auditory and olfactory) into spike trains, which can be used downstream for spike-based multi-modal learning tasks.

\subsection{Existing silicon retinas}\label{sec:event-based_vision}

The understanding that the retina deals with the inherent unreliability of `wetware' by responding to relative intensity changes (rather than the absolute magnitude) in the visual field \cite{Mead_1989_AdaptiveRetina} has been central to the development of silicon analogues to the retina. One of the earliest circuits that modelled the adaptive behaviour of retinal cells was a photoreceptor that mimicked some homeostatic processes in the retina. Specifically, the receptor could detect \textit{changes} in illumination rather than the absolute illumination level \cite{Mead_1985_SensitiveElectronicPhotoreceptor,DelbruckMead_1988_ElectronicPhotoreceptorSensitive}. The idea behind this circuit was inspired by the operation of receptors and bipolar cells (see \S \ref{par:retina.bipolar_cells}). It employed a feedback mechanism that compared the response to the changes in intensity to a filtered version of the output of the receptor, with an added hysteretic element that provided a `memory' of recent illumination. Importantly, the circuitry emulated the way that biological photoreceptors adapt to changes in light intensity on a logarithmic rather than linear scale (a phenomenon known as Weber's law). In other words, if the output of the receptor is given as $ V \propto log(I) $, then the change in that response would be logarithmic as well ($ \delta{V} \propto {\delta{I}}/{I} $).

A prominent milestone along this line of research was the development of a `silicon retina' \cite{MahowaldMead_1991_SiliconRetina}, where each photoreceptor could adapt to the local light intensity through feedback from simulated horizontal cells and the adapted signal was further amplified by silicon analogues of bipolar cells. The silicon retina demonstrated how a globally connected matrix of elements can still adapt to local fluctuations in light intensity, producing a familiar output comparable to that of an actual biological retina. Ultimately, a circuit was demonstrated that could model the behaviour of all five types of retinal cells, including ganglion cells \cite{ZaghloulBoahen_2006_SiliconRetinaThat}.

Subsequently, the Address Event Representation (AER) \cite{Serrano-GotarredonaAndreouLinares-Barranco_1999_AERImageFiltering,Serrano-GotarredonaSerrano-GotarredonaAcosta-JimenezEtAl_2006_NeuromorphicCorticalLayerMicrochip} was developed with a slightly different objective, namely efficient communication of a continuous signal in the form of multiplexed spike trains. The generic nature of AER was used to demonstrate how a circuit could perform spatial convolution (using, for instance, Gabor filters), allowing for feature extraction and tracking with high temporal resolution.

\subsubsection{The dynamic vision sensor (DVS)}

One of the main problems with the early circuits described above was that they were built primarily as proof-of-concept devices -- they were never meant to be used as actual camera sensors. This changed with the development of the dynamic vision sensor (DVS) \cite{BrandliBernerYangEtAl_2014_240180130,TaverniPaulMoeysLiEtAl_2018_FrontBackIlluminated}.

\begin{figure}[hpt]
    \centering
    \includegraphics[width=\textwidth]{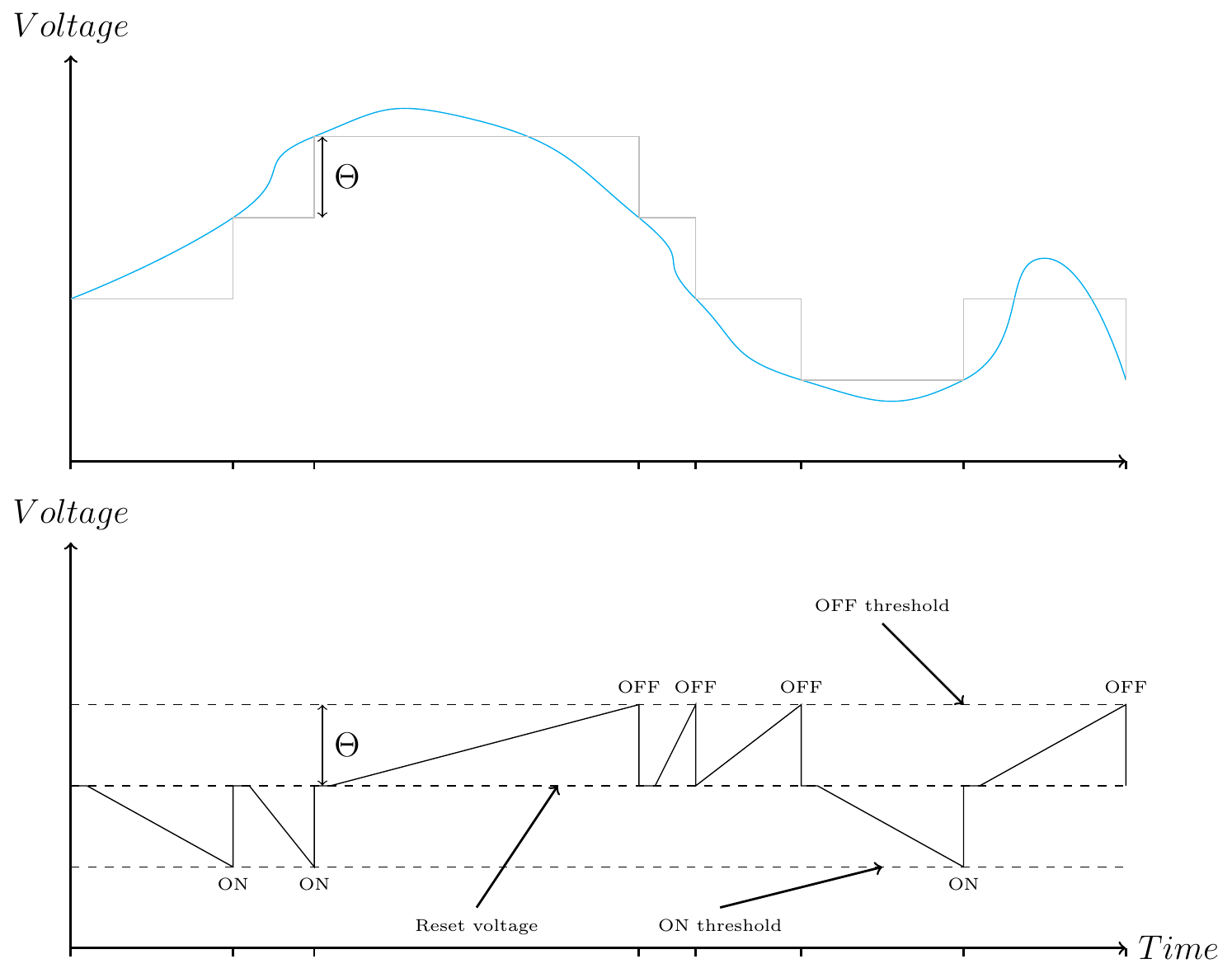}
    \caption{Operating principle of a DVS pixel. Every time the brightness of the area of the visual scene covered by the pixel increases (decreases) by a fixed amount (quantified by the threshold $\Theta$), the pixel generates an ON (OFF) event. The AER representation is commonly used for encoding the output of an event sensor, whereby an AER event consists of a timestamp, the $x$ and $y$ coordinates of the pixel that generated the event and the polarity of the output ($+1$ for ON and $-1$ for OFF events).} \label{fig:dvs.pixel}
\end{figure}

In many ways, DVS is a simplification on the original approach of modelling the retina holistically (i.e., all the layers separately). For instance, pixels in the DVS camera are entirely independent of all other pixels, allowing for a completely asynchronous operation that employs AER to output events ($ x $ and $ y $ coordinates, timestamp $ t $ and polarity $ p $). DVS pixels are relatively simple and implement the entire pipeline from receptors through bipolar cells and ganglion cells within a single circuit. The main principle of event generation and adaptation remains similar to that in earlier research: each pixel detects changes in the brightness ($ \log $ intensity) in either positive or negative direction relative to a baseline. Once the derivative of the logarithmic-intensity crosses a fixed threshold $ \Theta $, an event is produced and the brightness value in correspondence of the threshold plus some delay time set as the new baseline. In this way, rapid changes in brightness result in a large number of events, whereas slow fluctuations produce few events. Importantly, since the baseline is \textit{reset independently for each pixel} to the level of the threshold each time the respective pixel produces an event, the sensor as a whole has an exceedingly large dynamic range -- on the order of $140 dB$. This means that stark contrast does not saturate the sensor because the response of a pixel on the dark side of an edge does not depend on the response on the bright side of the same edge; it only depends on the \textit{relative} changes in brightness (Fig. \ref{fig:dvs.pixel}).


The DAVIS sensor \cite{BrandliBernerYangEtAl_2014_240180130} adds active pixel sensor circuitry to the core DVS logic in order to be able to keep information about the absolute brightness in the scene while retaining the advantages of the DVS sensor, such as high dynamic range, low latency and sparse activity. This allows a conventional frame-based readout to be obtained simultaneously with the fast asynchronous event-based output of the DVS sensor.

DVS pixels also show a characteristic junction leakage current, which depends strongly on the temperature \cite{NozakiDelbruck_2017_TemperatureParasiticPhotocurrent}. This leakage current is the source of sporadic (noisy) ON events, with the noise level becoming more prominent at high temperature and in dark environments.

The DVS pixel circuits enabled event-based sensors to bridge the gap between proof of concept and usability, leading to several commercial implementations (e.g., by Inivation\footnote{\url{https://inivation.com/}} and Prophesee\footnote{\url{https://www.prophesee.ai/}}).

There are also some notable points where the DVS departs from the operating principles of the mammalian retina. For instance, the DVS does not fully preserve the ON/OFF pathway separation since a single pixel can produce both ON and OFF events (whereas in biological systems these pathways are in fact kept separate all the way to the visual cortex \cite{MazadeJinPonsEtAl_2019_FunctionalSpecializationCortical}). Also, due to the lack of cross-talk between pixels, it is impossible for DVS cameras to implement the centre/surround RFs of bipolar and ganglion cells (although recent research aimed to introduce RFs \cite{DelbruckLiGracaEtAl_2022_UtilityFeasibilityCenter}, mainly with the goal of suppressing noise). Other features of the retina and the eye in general, such as microsaccades \cite{HajizadaBerggoldIaconoEtAl_2022_InteractiveContinualLearninga} and foveation \cite{GruelHarebGrimaldiEtAl_2022_StakesNeuromorphicFoveationa}, have also been attracting the attention of event-based vision researchers in recent years.
In parallel, there are also efforts to develop software that can emulate, simulate or otherwise complement the capabilities offered by hardware event-based sensors. For instance, the development of the video2events (v2e) framework \cite{Hu2021} and the ESIM library \cite{RebecqGehrigScaramuzza_2018_ESIMOpenEvent} was driven by the objective to harness existing conventional video datasets for event-based vision research.

\subsection{Why in space?}
In 2021, he DAVIS240 sensor became the first neuromorphic device to be launched in space and is, at the time of writing, also onboard the ISS launched as part of the ISS resupply payload in the Falcon Neuro Project \cite{mcharg2022falcon}. 
This shows the great interest aroused by neuromorphic sensing on the aerospace experts. The understanding and quantification of the advantages of event based cameras in the context of space missions is, though, only at its beginning \cite{Sikorski2021, McLeod2022, chin2019star, jawaid2022towards, ralph2022astrometric}. 
Besides the discussed benefits of pursuing a neuromorphic approach whenever low-resource requirements are driving the design of a mission, additional advantages over a convential camera  could derive from its higher dynamic range, achieving superior performance in high-contrast scenes as well as allowing a relaxation on the constrains imposed by light entering the field of view of the sensor.
The high temporal resolution and compression qualities of an event sensor are also of interest, as well as its recently proved resilience to different types of radiation (such as wide-spectrum neutrons) that are often present in space operating environments \cite{roffe2021neutron}.
Promising scenarios where event based cameras could make a difference in a space context are landing \cite{Sikorski2021, McLeod2022}, pose estimation \cite{jawaid2022towards}, planetary surface mapping, astronomy \cite{chin2019star, ralph2022astrometric}, monitoring of specific events (e.g., explosive events, pebble dynamics around asteroids, etc.). 
In all these cases it can be argued that the neuromorphic nature of the camera (combining high power efficiency and other beneficial properties), could bring advantages with respect to conventional systems, especially when a neuromorphic approach also used from the algorithmic side to tackle the interpretation of the produced events \cite{Sikorski2021}.  
However, an open question that remains in this field is the quantification of the domain gap introduced by the use of synthetic data produced by video-to-event converters (a classical example being v2e \cite{Hu2021}). A recent work \cite{jawaid2022towards} addresses this issue by proposing and testing an algorithm (trained on a synthetic dataset) on event streams created by a realistic mock-up experiment performed in a lab.

\subsection{Landing with events}
The ACT has recently demonstrated the suitability of event-based vision for autonomous planetary landing operations by inferring and processing optical flow measurements to predict time-to-contact (TTC) and divergence. TTC is defined as:
\begin{equation}
    \text{TTC} = - \frac{z}{v_z}
    \label{eq:TTC}
\end{equation}
where $z$ is the altitude of the spacecraft with respect to the planetary surface and $v_z$ its vertical velocity, so that a descending spacecraft will always satisfy $z>0$ and $v_z < 0$.
Previous work has already demonstrated the suitability of TTC feedback for controlling a spacecraft during flight operations and landing scenarios by using visual information from traditional frame-based sensors onboard micro air vehicles \cite{Izzo2012, Ho2017}. Our recent studies have shown that event data could also be effective for reconstructing on-board TTC, therefore bridging the gap from on-board event data streams to real-time control \cite{Sikorski2021, McLeod2022}.

In these studies, event streams representative of relevant ventral landing scenarios were constructed synthetically by converting sequences of frames similar to those provided by traditional frame-based sensors to events. The PANGU (Planet and Asteroid Natural Scene Simulation Utility)\cite{PANGU} software was used to generate proxy models of the Moon surface and render frames representative of the output of onboard vision instrumentation. From the generated synthetic events, TTC was estimated by identifying features in the event stream corresponding to static structures in the planetary body (such as craters and boulders) and computing their rate of perceived expansion or optical flow divergence. The reconstructed TTC from the event stream was fed back into a closed-loop control system to simulate real-time autonomous ventral landing by enforcing constantly decreasing TTC. Simulations were performed for different lunar terrains characterized by diverse crater and boulder distributions. Results showcased the suitability of the solution as a proof of concept for real-time onboard processing of events in planetary landing scenarios, where the main areas with room for improvement that could benefit from future research are the inference of divergence and the inclusion of more accurate noise models for event modelling based on known properties of DVS hardware.\cite{Sikorski2021}

In recent work, the divergence estimation procedure was further improved by using a contrast maximisation \cite{Gallego2018, Stoffregen2019} formulation for event-based divergence estimation, demonstrating its applicability on both synthetic and real event datasets.\cite{McLeod2022} Spacecraft position and velocity vectors were obtained for several planetary ventral landing profiles computed via indirect optimisation methods. A UR5 robotic arm with a Prophesee GEN 4 event sensor was employed to replicate these ventral trajectories using 2D and 3D printed planetary surfaces. Some of the generated trajectories were used to render 3D landing reconstructions in PANGU, which were then passed through the v2e \cite{Hu2021} pipeline to generate synthetic event streams. A mathematical formulation for the event-based radial flow was proposed, and a GPU-accelerated optimisation procedure was used to maximise the contrast of the resulting flow-compensated event images and estimate the divergence of event batches. Comparisons to other state-of-the-art divergence estimation methods showcased the accuracy and stability of the proposed procedure across a wide variety of event streams, with competitive run times achieved owing to the GPU-accelerated implementation.

\subsection{From events to egomotion}
Events as produced by the DVS or some other device based on retinomorphic models contain information on the perceived scene motion as projected in the camera plane. But what can be learned about the spacecraft egomotion (or pose) using only this information?
It is conceivable that events can be used for reconstructing the motion field, whereby the problem of motion field interpretation has already been discussed in the literature \cite{gupta19953}. 
A number of researchers have commented on this issue, starting with the work by Longuet Higgins \cite{longuet1980interpretation}, which addressed the question for the case of retinal vision, and subsequent works that lay down the mathematical structure of the problem, mapping it into a linear problem and thus marking it as \lq\lq  solved\rq\rq\ \cite{gupta19953}.
More recently, substantial work in the area of drones and robotic vision has been produced based on these fundamental results established in the 80s and 90s.

Let us consider the familiar equations \cite{longuet1980interpretation, gupta19953}:
$$
    \left\{
    \begin{array}{l}
        u =  (v_{xc} + xv_{zc}) h(x,y) + q+ry - pxy+ qx^2 \\
        v =  (v_{yc} + yv_{zc}) h(x,y) + rx-p - py^2+ qxy
    \end{array}
    \right.
$$
where $u,v$ are the velocities of the feature in $x,y$ on the camera plane, $p,q,r$ the angular velocity components of the camera, $v_{xc}, v_{yc}, v_{zc}$ the velocity of the camera center of mass and $h(x,y)$ the inverse of the depth map.
\begin{figure}[tb]
    \centering
    \includegraphics[width=0.9\linewidth]{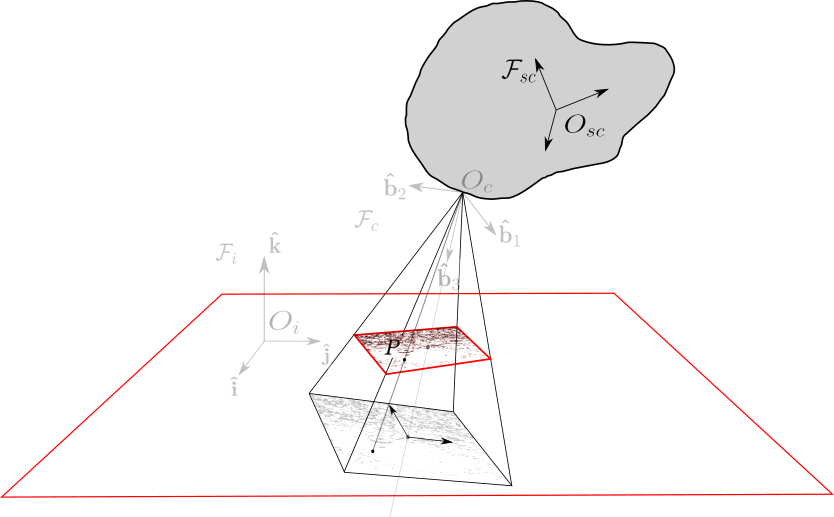}
    \caption{Camera geometry during planetary landing. The three reference systems, namely planet (i), camera (c) and spacecraft (sc), are shown as well as the camera plane projection of a generic surface feature assumed to be stationary. \label{fig:camera}}
\end{figure}
From these equations, if $h,v_{xc}, v_{yc}, v_{zc}$ constitutes a solution, so does $\frac hk, kv_{xc}, kv_{yc}, kv_{zc}$; it is thus impossible to invert the motion field to absolute values of all motion parameters, as only relative depth can be distinguished. Approximating the surface as an infinite plane, one can show that necessarily:
$$
h(x,y) = \alpha x + \beta y + \gamma
$$
where $H^2(\alpha^2 +\beta^2+\gamma^2) = 1$, having denoted the spacecraft altitude with $H$. Let us analyse the structure of the resulting equations. If we assume to know the motion field $u_i, v_i, \quad i= 1..n$ at $n$ different points, we can write a system of $2n$ equations in the unknowns $v_{xc}, v_{yc}, v_{zc}, p, q, r, \alpha, \beta, \gamma$. As before, only relative depth can be estimated, hence one can set $H=1$ and consider $\gamma^2 = 1 - \alpha^2 - \beta^2$ to conclude that in theory $n=4$ points of the motion field are necessary and sufficient to fully determine the spacecraft egomotion in relative terms to the altitude $H$. This conclusion does not consider the effect of measurement noise on the resulting estimates, a problem that is highly dependent on the sensor used (i.e., the quality of the events produced and their translation into a motion field). To conclude, it is worth mentioning the recent work from Jawaid \etal \cite{jawaid2022towards} on pose estimation from events. In a docking scenario (both simulated and reproduced in a lab mockup), the approach proposed in that work is able to determine from event data also the absolute values of mutual distances. This is not in contradiction to what is claimed above, as the algorithm is not exploiting depth cues from a reconstructed motion field but rather cues learned directly from a predefined dataset of known poses where the absolute dimensions appear explicitly.

\printbibliography


\end{document}